\documentclass[runningheads]{llncs}
\usepackage{graphicx}

\usepackage{tikz}
\usepackage{comment}
\usepackage{amsmath,amssymb} 
\usepackage{color}
\usepackage{booktabs}
\usepackage{multirow}
\usepackage{blindtext}
\usepackage{array}
\usepackage{bm}
\usepackage{wrapfig}
\usepackage{bbm}
\usepackage[hidelinks]{hyperref}

\usepackage{pifont}
\newcommand{\cmark}{\ding{51}}%
\newcommand{\xmark}{\ding{55}}%

\newcommand{\PreserveBackslash}[1]{\let\temp=\\#1\let\\=\temp}
\newcolumntype{C}[1]{>{\PreserveBackslash\centering}p{#1}}

\begin{document}
\pagestyle{headings}
\mainmatter
\def\ECCVSubNumber{3378}  

\title{Gen6D: Generalizable Model-Free 6-DoF Object Pose Estimation from RGB Images} 

\titlerunning{Gen6D Pose Estimator}
%
\author{Yuan Liu\inst{1} \and
Yilin Wen\inst{1} \and
Sida Peng\inst{2} \and
Cheng Lin\inst{3} \and \\
Xiaoxiao Long\inst{1} \and
Taku Komura\inst{1} \and
Wenping Wang\inst{4}
}
\authorrunning{Y. Liu, Y. Wen, S. Peng, et al.}
%
\institute{The University of Hong Kong \and
Zhejiang University \and
Tencent \and
Texas A\&M University
}
\maketitle

\begin{abstract}
In this paper, we present a generalizable model-free 6-DoF object pose estimator called Gen6D. Existing generalizable pose estimators either need the high-quality object models or require additional depth maps or object masks in test time, which significantly limits their application scope. In contrast, our pose estimator only requires some posed images of the unseen object and is able to accurately predict poses of the object in arbitrary environments. Gen6D consists of an object detector, a viewpoint selector and a pose refiner, all of which do not require the 3D object model and can generalize to unseen objects. Experiments show that Gen6D achieves state-of-the-art results on two model-free datasets: the MOPED dataset and a new GenMOP dataset. In addition, on the LINEMOD dataset, Gen6D achieves competitive results compared with instance-specific pose estimators. Project page: \href{https://liuyuan-pal.github.io/Gen6D/}{https://liuyuan-pal.github.io/Gen6D/}.
\end{abstract}

\section{Introduction}

Estimating the orientation and location of an object in 3D space is a preliminary and necessary step for many tasks involving interaction with the object. In the last decade, 3D vision has witnessed tremendous development ranging from robotics, games, to VR/AR. These applications raise new demands for the 6-DoF object pose estimation, requiring a pose estimator to be generalizable, flexible, and easy-to-use. However, existing methods suffer from several restrictive conditions. Most methods~\cite{labbe2020cosypose,xiang2017posecnn,wang2019normalized} can only be used for a specific object or category same as the training data. Some methods~\cite{sundermeyer2020multi,li2018deepim,zakharov2019dpod,xiao2019pose,pitteri20203d,park2020latentfusion,yen2021inerf} can generalize to unseen objects, but they rely on high-quality target 3D models~\cite{sundermeyer2020multi,li2018deepim,zakharov2019dpod,xiao2019pose,pitteri20203d}, or additional depth maps~\cite{park2020latentfusion} and masks~\cite{park2020latentfusion,yen2021inerf} at test time. These requirements severely limit the practical applications of the existing pose estimators.

\begin{figure}[htbp]
    \centering
    \includegraphics[width=\textwidth]{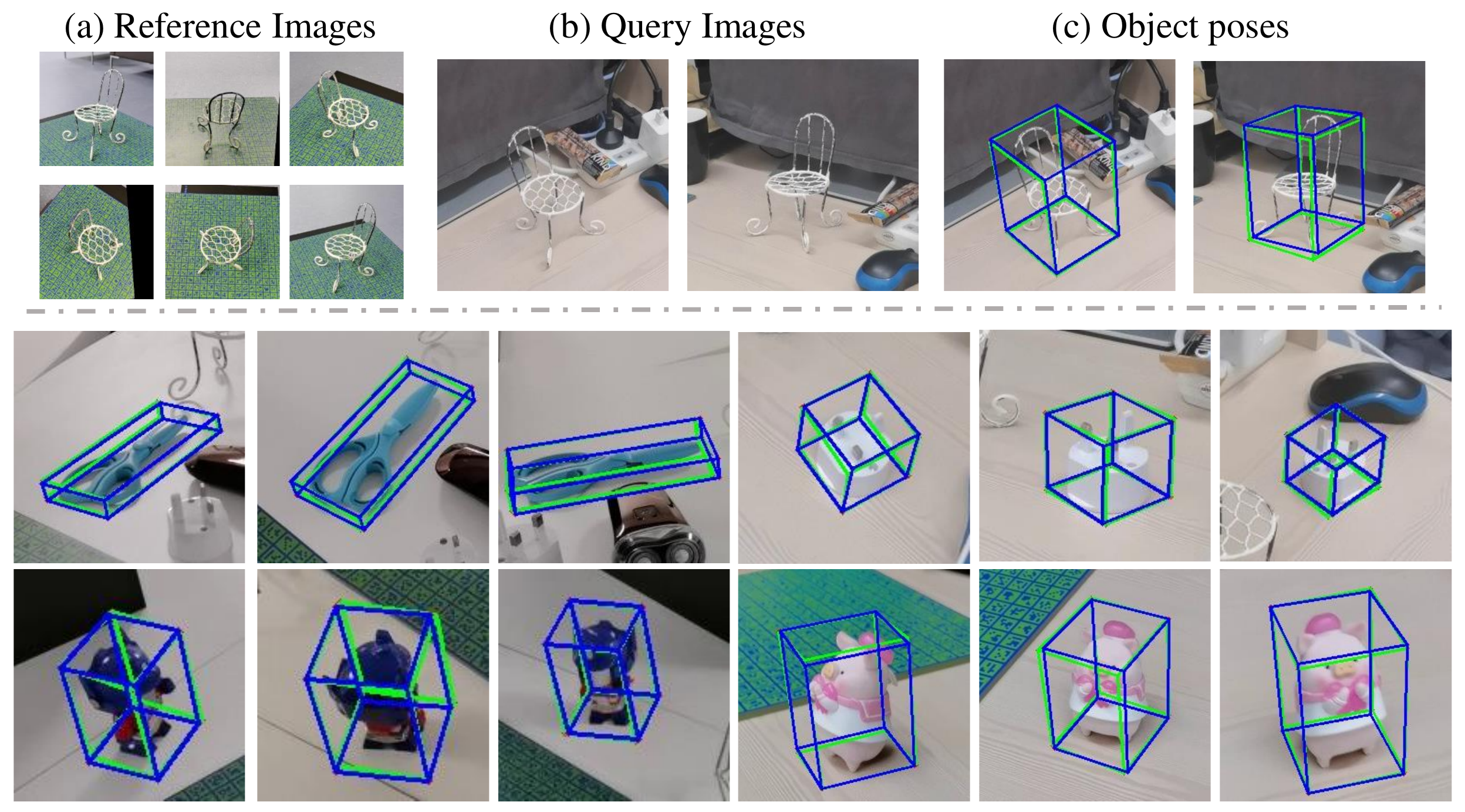}
    \caption{Given (a) reference images of an object with known poses and (b) query images containing the same object with unknown poses, our pose estimator is able to accurately estimate (c) their object poses in the query images, where {green color} means ground-truth and {blue color} means estimation. \textbf{Note that all objects are unseen in the training set and the same estimator is applied for all objects}.}
    \label{fig:teaser}
\end{figure}

To meet demands in practical applications, we argue that such a pose estimator should have the following properties.
1) \textbf{Generalizable}. The pose estimator can be applied on an arbitrary object without training on the object or its category. 
2) \textbf{Model-free}. When generalizing to an unseen object, the estimator only needs some reference images of this object with known poses to define the object reference coordinate system, as shown in Fig.~\ref{fig:teaser} (a), but does not rely on a 3D model of the object.
3) \textbf{Simple inputs}. When estimating object poses, the estimator only takes RGB images as inputs without requiring additional object masks or depth maps.



\begin{figure}
    \centering
    \includegraphics[width=\textwidth]{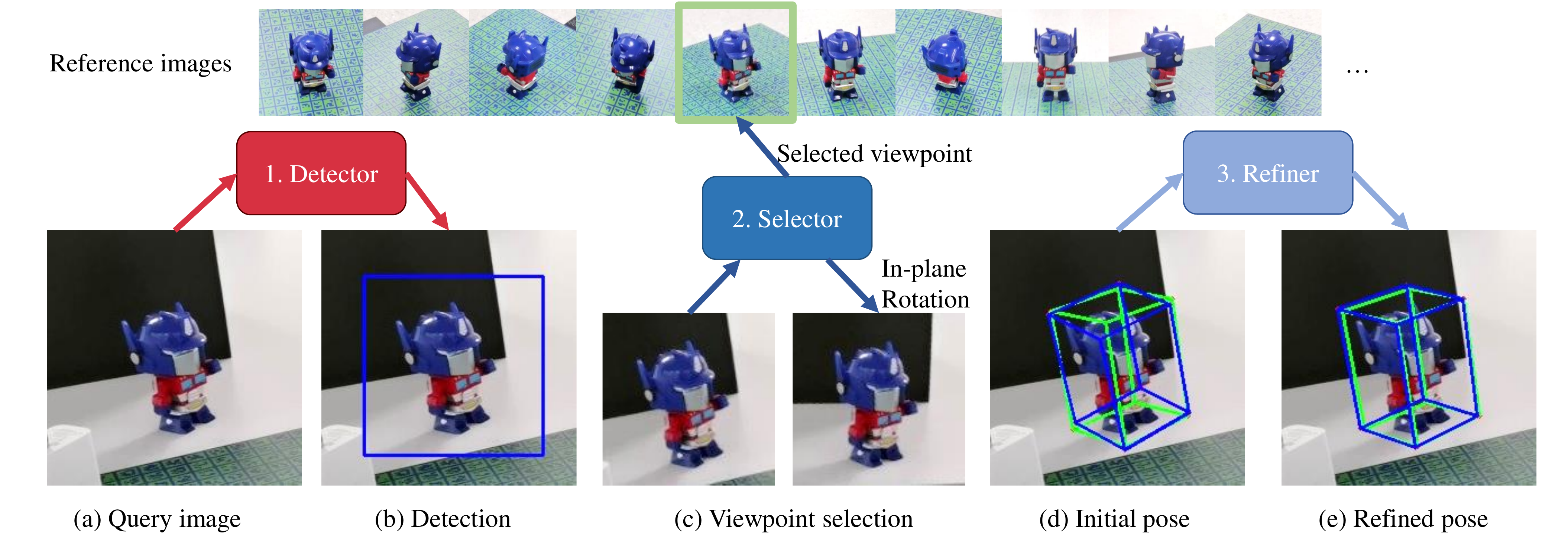}
    \caption{\textbf{Overview}. The proposed pose estimator consists of a detector which detects the object in the query image, a viewpoint selector which selects the most similar viewpoint from reference images, and a pose refiner which refines the initial pose into an accurate object pose.}
    \label{fig:overview}
\end{figure}

To the best of our knowledge, there is no existing pose estimator satisfying all the above three properties simultaneously. Thus, in this paper, we propose a simple but effective pose estimator, called \textbf{Gen6D}, which possesses the three properties above. 
Given input reference images of an arbitrary object with known poses, Gen6D is able to directly predict its object pose in any query images, as shown in Fig.~\ref{fig:teaser}.
In general, an object pose can be estimated by directly predicting rotation/translation by regression~\cite{xiang2017posecnn,hu2020single,tekin2018real}, solving a Perspective-n-Points (PnP) problem~\cite{peng2019pvnet,rad2017bb8} or matching images with known poses~\cite{sundermeyer2018implicit,wohlhart2015learning,sundermeyer2020multi}. 
Direct prediction of rotation and translation by regression is mostly limited to a specific instance or category, which has difficulty in generalizing to unseen objects.
Meanwhile, due to the lack of 3D models, PnP-based methods do not have 3D keypoints to build 2D-3D correspondences so that they are incompatible with model-free setting. 
Hence, we apply image-matching in our framework for pose estimation, which can  generalize to unseen objects by learning a general image similarity metric. 

In Gen6D, we propose a novel image-matching based framework to estimate the object pose in a coarse-to-fine manner. The framework consists of an object detector, a viewpoint selector and a pose refiner, as shown in Fig.~\ref{fig:overview}. 
Given reference images and a query image, an object detector first detects the object regions by correlating the reference images with the query image, which is similar to \cite{ammirato2018target}.
Then, a viewpoint selector matches the query image against the reference images to produce a coarse initial pose.
Finally, the initial pose is further refined by a pose refiner to search for an accurate object pose.

\begin{figure}
  \centering
  \begin{tabular}{ccc|ccc}
       Query & Nearest & 2nd Nearest & Query & Nearest & 2nd Nearest \\
       \includegraphics[width=0.15\textwidth]{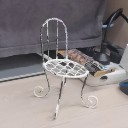} &  
       \includegraphics[width=0.15\textwidth]{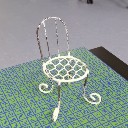} &  
       \includegraphics[width=0.15\textwidth]{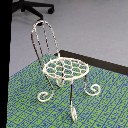} &
       \includegraphics[width=0.15\textwidth]{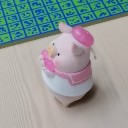} &  
       \includegraphics[width=0.15\textwidth]{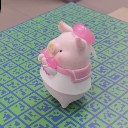} &  
       \includegraphics[width=0.15\textwidth]{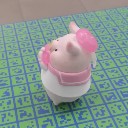}
  \end{tabular}
  
  \caption{Query images and reference images have cluttered background. The reference image with the nearest viewpoint looks very similar as the one with second nearest viewpoint, which brings challenges for the selector to correctly select the nearest one.}
  \label{fig:imp}
\end{figure}

A challenge is how to design a viewpoint selector when the reference images are sparse and contain cluttered background. Existing image-matching methods~\cite{sundermeyer2018implicit,wohlhart2015learning,sundermeyer2020multi,hinterstoisser2011multimodal,balntas2017pose} have difficulty in handling this problem due to two problems. First, these image-matching methods embed images into feature vectors and compute similarities using distances of feature vectors, in which cluttered background interferes the embedded feature vectors and thus severely degenerates the accuracy. 
Second, given a query image, there may not be a reference image with exactly the same viewpoint as the query image. In this case, there will be multiple plausible reference images and the selector has to select the one with the nearest viewpoint as the query image, which usually are very ambiguous as shown in Fig.~\ref{fig:imp}.

To address these problems in viewpoint selection, we propose to use neural networks to pixel-wisely compare the query image with every reference image to produce similarity scores and select the reference image with highest similarity score. This pixel-wise comparison enables our selector to concentrate on object regions and reduces the influence of cluttered background. 
Furthermore, we add global normalization layers and self-attention layers to share similarity information across different reference images. 
These two kinds of layers enable every reference images to commute with each other, which provides context information for the selector to select the most similar reference image.

  

The main challenge of developing our pose refiner is the unavailability of the object model.
Existing pose refiners~\cite{li2018deepim,zakharov2019dpod} are based on rendering-and-comparison, which render an image on the input pose and then match the rendered image with the query image to refine the input pose. However, without the object model, rendering high-quality images on arbitrary poses is difficult, which makes these refinement methods infeasible in the model-free setting. 

To address this problem, we propose a novel 3D volume-based pose refinement method. Given a query image and an input pose, we find several reference images that are near to the input pose. These reference images are projected back into 3D space to construct a feature volume. Then, the constructed feature volume is matched against the features projected from the query image by a 3D CNN to refine the input pose. In comparison with previous pose refiners~\cite{li2018deepim,zakharov2019dpod}, our pose refiner avoids rendering any new images. Meanwhile, the constructed 3D feature volume enables our method to infer the 3D pose refinement in the 3D space. In contrast, previous pose refiners~\cite{li2018deepim,zakharov2019dpod} only rely on 2D image features to regress a 3D relative pose, which are less accurate especially for unseen objects.

To validate the effectiveness of our generalizable model-free pose estimator, we introduce a new dataset, called General Model-free Object Pose Dataset (GenMOP), which contains video sequences of objects in different environments and lighting conditions. We choose one sequence as reference images and the rest sequences of the same object as test query images. Experiments show that without training on these objects, our method still outperforms instance-specific estimator PVNet~\cite{peng2019pvnet} on the GenMOP dataset and another model-free MOPED~\cite{park2020latentfusion} dataset. We also evaluate our method on the LINEMOD dataset~\cite{hinterstoisser2012linemod}, on which our generalizable pose estimator achieves comparable results as instance-specific estimators which needs to be trained with excessive rendered images. 

\section{Related works}


\subsection{Specific object pose estimator}
Most object pose estimators~\cite{xiang2017posecnn,sundermeyer2018implicit,peng2019pvnet,labbe2020cosypose,hodan2020epos,wen2020edge,di2021so,wang2021gdr,liu2020keypose,song2020hybridpose,hu2020single,pitteri2019object,hodavn2020bop,hodan2018bop,ponimatkin2022focal,su2022zebrapose} are instance-specific, which cannot generalize to unseen objects and usually require a 3D model of the object to render extensive images for training. 
Recent instance-specific pose estimators~\cite{park2021dprost,cai2020reconstruct,liu2021stereobj} reconstruct the object model implicitly in the pipeline so that they are model-free. Category-specific pose estimators~\cite{wang2019normalized,chen2020category,wen2021disentangled,deng2022icaps,lin2021sparse,chen2021fs,tian2020shape,chen2020learning,lin2021dualposenet,chen2021sgpa,goodwin2022zero,di2022gpv} can generalize to objects in the same category and also do not require the object model. However, they are still unable to predict poses for objects in unseen categories. 
In comparison, Gen6D is  generalizable, which makes no assumption of the category or the instance of the object and also does not need the 3D model of the object. 

\subsection{Generalizable object pose estimator}
Generalizable pose estimators mostly require an object model either for shape embedding~\cite{xiao2019pose,pitteri20203d,dani20213dposelite,pitteri2019cornet} or template matching~\cite{hinterstoisser2011multimodal,balntas2017pose,wohlhart2015learning,sundermeyer2020multi,hinterstoisser2011gradient,nguyen2022templates,zhao2022fusing} or rendering-and-comparison~\cite{li2018deepim,zakharov2019dpod,okorn2021zephyr,busam2020like,sundermeyer2020multi,grabner2020geometric}. To avoid using 3D models, recent works~\cite{yen2021inerf,park2020latentfusion} utilize the advanced neural rendering techniques~\cite{mildenhall2020nerf} to directly render from posed images for pose estimation. 
However, current rendering methods are only able to render images under exactly the same appearance like lighting conditions, which degenerates the accuracy under varying appearance. To remedy this, these methods have to resort to additional depth maps~\cite{park2020latentfusion} or object masks~\cite{park2020latentfusion,yen2021inerf} to achieve robustness. There are also some works focusing on estimating poses of unseen objects using RGBD sequences~\cite{wen2021bundletrack,okorn2021zephyr,simeonov2021neural,he2022fs6d,cai2022ove6d,gou2022unseen}. In contrast to these methods, Gen6D is model-free and does not require depth maps or masks. There are also concurrent works~\cite{sun2022onepose,shugurov2022osop} 
of generalizable pose estimation.

\subsection{Instance detection}
Instance detection aims to detect a given object with some images of the object~\cite{ammirato2018target,mercier2021deep,gu2022ossid,osokin2020os2d}. There are some instance detection methods which also estimate viewpoints~\cite{xiao2020few,banani2020novel} for novel category in one- or few-shot setting. The detector of Gen6D is inspired by \cite{ammirato2018target}, which uses correlation to find the object region. The target of Gen6D is to estimate the 6-DoF object pose, which is different from these methods for detection or category-level viewpoint estimation.

\section{Method}

Given $N_r$ images of an object with known camera poses, called \textbf{reference images}, our target is to predict the pose of the object in a \textbf{query image}. The object pose here means a translation ${\bf t}$ and a rotation ${\bf R}$ that transform the object coordinate ${\bf x}_{obj}$ to the camera coordinate ${\bf x}_{cam} = {\bf R} {\bf x}_{obj} + {\bf t}$. All the intrinsics parameters of images are already known.

\textbf{Data normalization}. For every object, we can estimate a rough size of the object by triangulating points from reference images or simply unprojecting reference images to find an intersection. The center of triangulated points or the center of the 3D intersection region is regarded as the object center. Then, the object coordinate system is normalized so that the object center locates at the origin and the object size is 1, which means the whole object resides inside a unit sphere at the origin. 
This data normalization ensures the feature volume constructed by our pose refiner in Sec.~\ref{sec:ref} will contain the target object. 
More details about the normalization can be found in the supplementary materials.

\begin{figure}[htbp]
  \centering
  \includegraphics[width=0.95\textwidth]{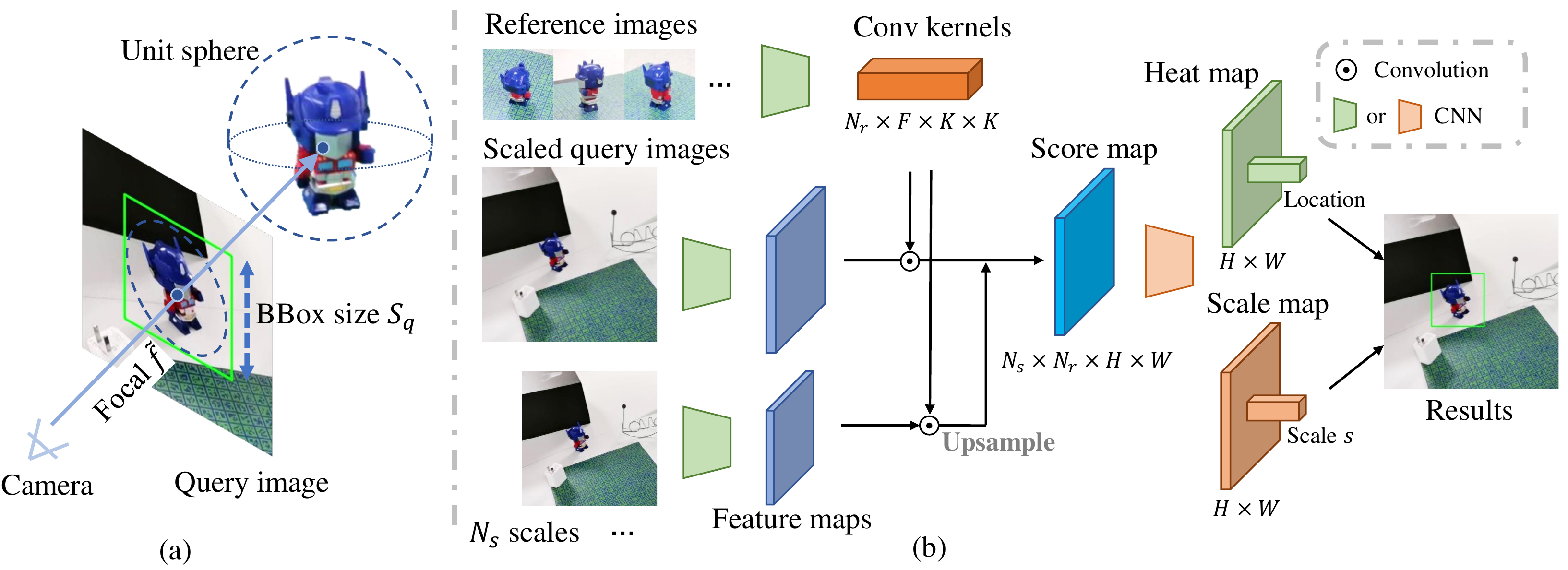}
  \caption{(a) Detection outputs. Depth can be computed from the bounding box size $S_q$, which along with the 2D projection of the object center determine the location of the object center. (b) Architecture of the detector. We use features of reference images as kernels to convolve features of multi-scale query images to get score maps. Score maps are further processed by a CNN to produce a heat map about the object center and a scale map to determine the bounding box size.}
  \label{fig:det}
\end{figure}

\textbf{Overview}. As shown in Fig.~\ref{fig:overview}, the proposed Gen6D pose estimator consists of an object detector, a view selector and a pose refiner.
The object detector crops the object region and estimates an initial translation (Sec.~\ref{sec:det}). The view selector finds an initial rotation by selecting the most similar reference image and estimating an in-plane rotation (Sec.~\ref{sec:sel}). The initial translation and rotation are used in the pose refiner to iteratively estimate an accurate pose (Sec.~\ref{sec:ref}).

\subsection{Detection}
\label{sec:det}

The query image is usually very large and the object only occupies a small region on the query image. 
To focus on the object, we apply a correlation-based instance detector similar to \cite{ammirato2018target}.
We decompose the detection problem into two parts, i.e. finding the 2D projection $q$ of the object center and estimating a compact square bounding box size $S_q$ that encloses the unit sphere.
As shown in Fig.~\ref{fig:det} (a), such a compact bounding box size is used in computing the depth of the object center by $d=2\tilde{f}/S_q$, where 2 is the diameter of the unit sphere and $\tilde{f}$ is a virtual focal length by changing the principle point to the estimated projection $q$.
The projection $q$ and the depth $d$ will determine the location of the object center, which provides an initial translation for the object pose.

\begin{figure}
    \centering
    \includegraphics[width=0.95\textwidth]{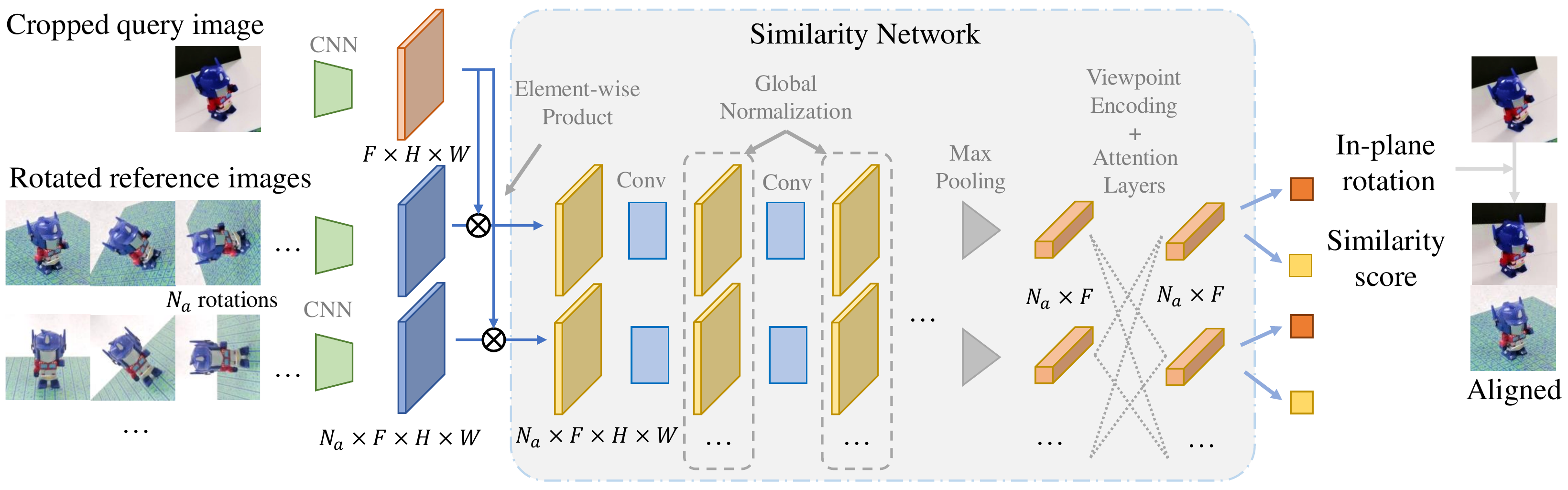}
    \caption{Architecture of the viewpoint selector. We compute the element-wise product of every reference image with the query image to get a score map, on which a similarity network is applied to compute an in-plane rotation and a similarity score for this reference image. Note that in the similarity network, we use global normalization layers and a transformer to share information across reference images.}
    \label{fig:sel}
\end{figure}

The design of our detector is shown in Fig.~\ref{fig:det} (b). We extract feature maps on both the reference images and the query image by a VGG~\cite{simonyan2014vgg}-11 network. Then, the feature maps of all reference images are regarded as convolution kernels to convolve with the feature map of the query image to get score maps. To account for scale differences, we conduct such convolution at $N_s$ predefined scales by resizing the query images to different scales. Based on the multi-scale score maps, we regress a heat map and a scale map. We select the location with the max value on the heat map as the 2D projection of the object center and use the scale value $s$ at the same location on the scale map to compute the bounding box size $S_q = S_r * s$, where $S_r$ is the size of reference images. 

With the detected 2D projections and scales, we compute the initial 3D translations and crop the object regions for subsequent processing. More details about architecture and training of detector networks can be found in the supplementary materials.

\subsection{Viewpoint Selection}
\label{sec:sel}

Viewpoint selection aims to select a reference image whose viewpoint is the nearest to the query image. Meanwhile, we will estimate an in-plane rotation between the query image and the selected reference image. We approximately regard the viewpoint of the selected reference image as the viewpoint of the query image, which along with the estimated in-plane rotation forms an initial rotation for the object pose.

As shown in Fig.~\ref{fig:sel}, we design a viewpoint selector to compare the query image with every reference image to compute similarity scores. Specifically, we first extract feature maps by applying a VGG~\cite{simonyan2014vgg}-11 on reference images and the query image. Then, for every feature map of reference images, we compute its element-wise product with the feature map of the query image to produce a correlation score map. Finally, the correlation score map is processed by a similarity network to produce a similarity score and a relative in-plane rotation to align the query image with the reference image. In our viewpoint selector, we have three special designs. 

\textbf{In-plane rotation}. To account for in-plane rotations, every reference image is rotated by $N_a$ predefined angles and all rotated versions are used in the element-wise product with the query image.

\textbf{Global normalization}. For every feature map produced by the similarity network, we normalize it with the mean and variance computed from all feature maps of reference images. Such a global normalization helps our selector select the relatively most similar reference image because it allows the distribution of feature maps to encode the context similarity and amplifies the similarity differences among different reference images. 
For every reference image, max-pooling is applied on its feature map to produce a similarity feature vector.

\textbf{Reference view transformer}. We apply a transformer on the similarity feature vectors of all reference images, which includes the positional encoding of their viewpoints and attention layers over all similarity feature vectors.
Such a transformer lets feature vectors communicate with each other to encode contextual information~\cite{vaswani2017attention,sarlin2020superglue,wang2021ibrnet}, which is helpful to determine the most similar reference image.
The outputs of reference view transformer will be used to regress a similarity score and an in-plane rotation angle for each reference image. The viewpoint of the reference image with highest score will be selected.

With the selected viewpoint and the estimated in-plane rotation, we estimated an initial rotation for the object pose, which will be refined by the pose refiner. More details about the network and training can be found in the supplementary materials.

\subsection{Pose refinement}
\label{sec:ref}

Combining the translation estimated by the object detector and the rotation estimated by the viewpoint selector, we get an initial coarse object pose. This initial pose is further refined by a 3D volume-based pose refiner.

\begin{figure}
    \centering
    \includegraphics[width=\textwidth]{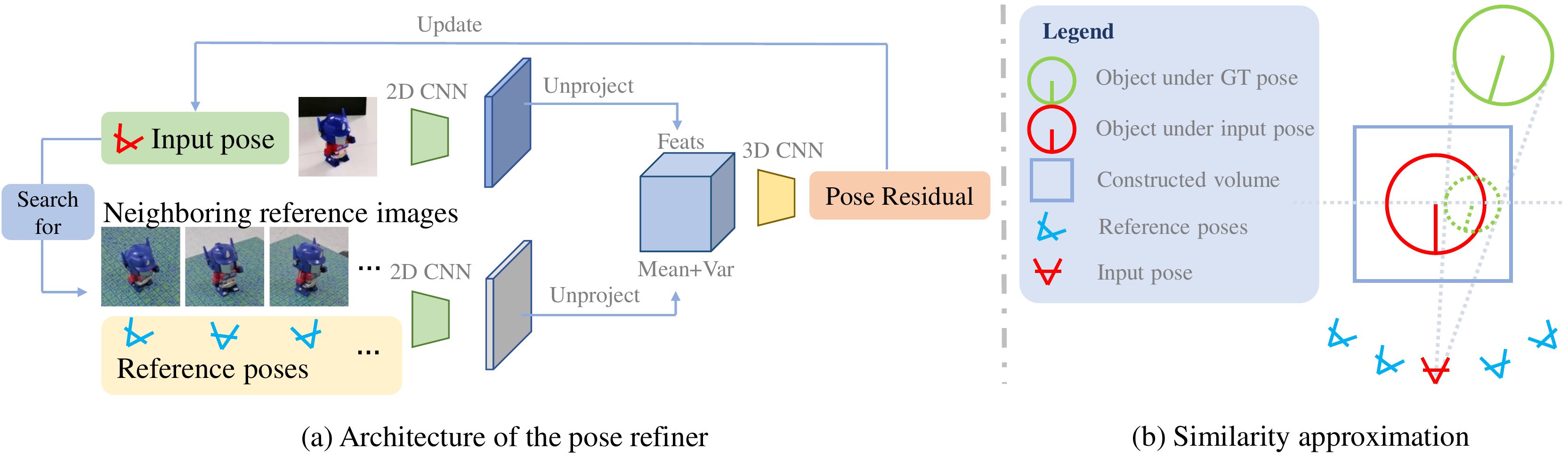}
    \caption{(a) Architecture of our pose refiner. (b) A 2D diagram to illustrate the similarity transformation approximation. Though the ground-truth pose residual from the input object pose (solid red circle) to the ground-truth object pose (solid green circle) is a rigid transformation, we can approximate this rigid transformation by a similarity transformation inside the feature volume. 
    Our pose refiner predicts the similarity transformation, which transforms the input red circle to the dotted green circle. 
    Then, the similarity transformation can be converted to a rigid transformation. 
    }
    \label{fig:ref}
\end{figure}

Specifically, since the objects are already normalized inside an unit sphere at the origin, we build a volume within the unit cube at the origin with $S^3_v=32^3$ vertices. As shown in Fig.~\ref{fig:ref} (a), to construct the features on these vertices, we first select $N_n=6$ reference images that are near to the input pose. We extract feature maps on these selected reference images by a 2D CNN. Then, these feature maps are unprojected into the 3D volume and we compute the mean and variance of features among all reference images as features for volume vertices. For the query image, we also extract its feature map by the same 2D CNN, unproject feature map into the 3D volume using the input pose and concatenate the unprojected query features with the mean and variance of reference image features. Finally, we apply a 3D CNN on the concatenated features of the volume to predict a pose residual to update the input pose.

\textbf{Similarity approximation}. Instead of regressing the rigid pose residual directly, we approximate it with a similarity transformation, as shown in Fig.~\ref{fig:ref} (b). The approximate similarity transformation consists of a 2D in-plane offset, a scale factor and a residual 3D rotation. The reason of using this approximation is that it avoids direct regression of the 3D translation from the red circle to the solid green circle in Fig.~\ref{fig:ref}, which is out of the scope of the feature volume. Instead, we regress a similarity transformation from red circle to dotted green circle, which can be easily inferred from the features defined in the volume. More details can be found in the supplementary materials. In our implementation, we apply the refiner iteratively 3 times by default.

\textbf{Discussion}. The key difference between our volume-based refiner and other pose refiners~\cite{li2018deepim,zakharov2019dpod,sundermeyer2020multi} is that our pose refiner does not require rendering an image on the input pose, which thus is more suitable for the model-free pose estimation. 
Meanwhile, since the 3D volume is constructed by multiple reference images with different poses, our volume-based refiner is able to know the image features under different poses and infer how pose changes affect the image features for unseen objects. 
In comparison, previous pose refiners~\cite{li2018deepim,zakharov2019dpod,sundermeyer2020multi} only compare a rendered image with the input query image to compute a pose residual. Such a 2D image does not provide enough 3D structure information to infer how pose changes affect image patterns, especially for unseen objects. Thus, it is hard for these methods to predict correct pose residuals for unseen objects.

\section{Experiments}

\subsection{GenMOP Dataset}
\begin{figure}
    \centering
    \setlength\tabcolsep{1.5pt} 
    \begin{tabular}{cccccccccc}
    \includegraphics[width=0.09\textwidth]{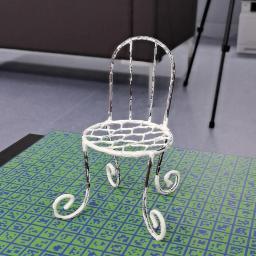} &
    \includegraphics[width=0.09\textwidth]{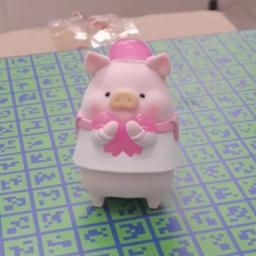} &
    \includegraphics[width=0.09\textwidth]{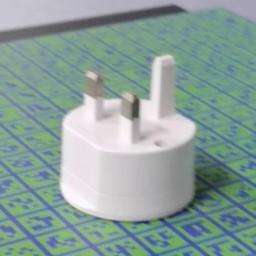} &
    \includegraphics[width=0.09\textwidth]{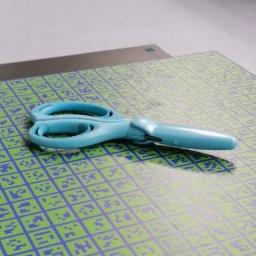} &
    \includegraphics[width=0.09\textwidth]{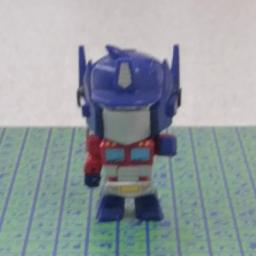} &
    \includegraphics[width=0.09\textwidth]{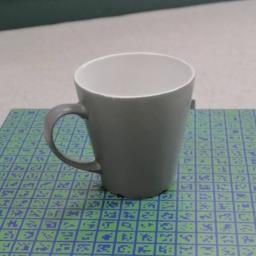} &
    \includegraphics[width=0.09\textwidth]{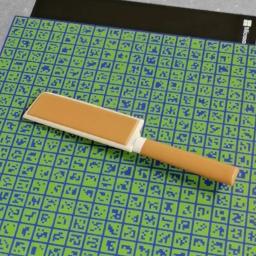} &
    \includegraphics[width=0.09\textwidth]{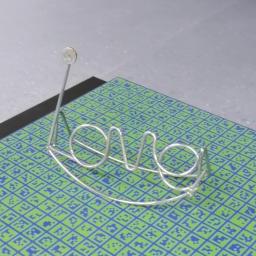} &
    \includegraphics[width=0.09\textwidth]{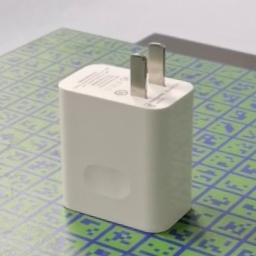} &
    \includegraphics[width=0.09\textwidth]{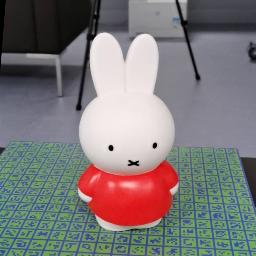} \\
    Chair& Piggy & PlugEN & Scissors & TFormer & Cup & Knife & Love & PlugCN & Miffy
    \end{tabular}
    \caption{Objects in the GenMOP dataset. The first 5 objects are used in test and the last 5 objects are used in training.}
    \label{fig:obj}
\end{figure}

To validate the effectiveness of the proposed method, we collect a dataset called General Model-free Object Pose Dataset (GenMOP). GenMOP dataset consists of 10 objects ranging from flatten objects like ``scissors" to thin structure objects like ``chair" as shown in Fig.~\ref{fig:obj}. For each object, two video sequences of the same object are collected in different environments like backgrounds and lighting conditions. Every video sequence is split into $\sim$200 images. For each sequence, we apply COLMAP~\cite{schoenberger2016colmap} to reconstruct the camera poses in each sequences separately and manually label keypoints on the object for cross-sequence alignment. More details about the GenMOP dataset can be found in the supplementary.

\subsection{Protocol}
We evaluate Gen6D pose estimator on the GenMOP dataset, the LINEMOD~\cite{hinterstoisser2012linemod} dataset and the MOPED~\cite{park2020latentfusion} dataset. 

\textbf{GenMOP}. On the GenMOP dataset, we select one video sequence as reference images and the other video sequence in a different environment as test query images, both of which contain $\sim$200 images. 

\textbf{LINEMOD}. The LINEMOD~\cite{hinterstoisser2012linemod} dataset is a widely-used dataset for object pose estimation. On the LINEMOD dataset, we follow the commonly-used train-test split as \cite{tekin2018real}. We select the training images ($\sim$180) as reference images and all the rest $\sim$1000 test images as query images for evaluation. 

\textbf{MOPED}. The MOPED~\cite{park2020latentfusion} dataset is intended for model-free object pose estimation. Since the MOPED dataset is generated automatically by depth fusion and point cloud registration, object poses in some sequences are not very accurate. Thus, we manually select reliable subsets from 5 objects for evaluation. For each object, there are 200-600 reference images and 100-300 query images.

\textbf{Training datasets}. The training dataset of Gen6D estimator consists of: 1) Rendered images from $\sim$2000 ShapeNet~\cite{chang2015shapenet} models, 2) Google Scanned Object dataset rendered by \cite{wang2021ibrnet} with 1023 objects, 3) 5 objects from the GenMOP dataset and 4) 5 objects (ape/can/holepuncher/iron/phone) from the LINEMOD dataset. Note we only train a single model and test its performance on the unseen objects on GenMOP, LINEMOD and MOPED dataset.

\textbf{Metrics}. We adopt the widely-used Average Distance (ADD)~\cite{hinterstoisser2012linemod} and the projection error as metrics. On the ADD, we compute the recall rate with 10\% of the object diameter (ADD-0.1d) and the AUC in 0-10cm (ADD-AUC). On the projection error, we compute the recall rate at 5 pixels (Prj-5).

\begin{table}[]
    \centering
    \caption{Performance on the GenMOP dataset. ``General" means generalizable or not. ``Ours w/o Ref." means not using the pose refiner in the Gen6D estimator.}
    \begin{tabular}{C{0.15\linewidth} lccccccc}
    \toprule
    \multirow{2}{*}{Metrics} & \multirow{2}{*}{Method} &  
    \multirow{2}{*}{General} &
    \multicolumn{5}{c}{Object Name} & \multirow{2}{*}{avg.}\\
    & & & Chair & PlugEN & Piggy & Scissors & TFormer & \\
    \midrule
    \multirow{5}{*}{ADD-0.1d}
    &PVNet~\cite{peng2019pvnet}     &   \xmark &
    49.50&  2.33&  \textbf{77.89}& \textbf{44.40}& 19.84& 38.79\\
    &RLLG~\cite{cai2020reconstruct}      & \xmark &  
    0.70 & 1.28 &1.01 &  3.45&  0.79& 2.71\\
    & ObjDesc~\cite{wohlhart2015learning}  &   \cmark &
    3.50& 5.14& 14.07& 1.25& 7.54& 8.55\\
    &Ours w/o Ref. &\cmark &
    14.00& 7.48&39.70& 16.81& 11.51& 17.90\\
    &Ours &\cmark &
    \textbf{61.50}& \textbf{19.63}& 75.38& 32.76&\textbf{62.70}& \textbf{50.39}\\
    \midrule
    \multirow{5}{*}{Prj-5}
    &PVNet~\cite{peng2019pvnet}     &   \xmark &
    15.00&  30.37&  83.42& \textbf{96.55}& 59.52& 56.97\\
    &RLLG~\cite{cai2020reconstruct}     &   \xmark &
    2.00&  4.67&  17.59& 35.78& 7.94& 13.59\\
    &ObjDesc~\cite{wohlhart2015learning}     &   \cmark &
    4.00&  10.75&  4.52& 18.53& 8.33& 9.23\\
    &Ours w/o Ref. & \cmark &
    11.50& 40.65& 33.17& 34.05& 64.29& 36.73\\
    &Ours &\cmark &
    \textbf{55.00}& \textbf{72.90}& \textbf{92.96}& 93.53&\textbf{98.81}& \textbf{82.64}\\
    \bottomrule
    \end{tabular}
    \label{tab:genmop}
\end{table}

\subsection{Results on GenMOP}
For comparison, we choose the generalizable image-matching based ObjDesc~\cite{wohlhart2015learning} and two instance-specific estimators PVNet~\cite{peng2019pvnet} and RLLG~\cite{cai2020reconstruct} as baseline methods. Quantitative results are shown in Table~\ref{tab:genmop} and some qualitative results are shown in Fig.~\ref{fig:teaser}. More qualitative results are in the supplementary.

\textbf{Baseline implementation}. 
For the generalizable template-matching ObjDesc~\cite{wohlhart2015learning}, we use the same training dataset as Gen6D. In testing, we crop the object region by our object detector and then use ObjDesc to select the most similar reference image to the query image. The pose of the selected reference image is regarded as the pose of the query image. 
All objects used in evaluation are unseen for Gen6D and ObjDesc in training.
For instance-specific estimators PVNet~\cite{peng2019pvnet} and RLLG~\cite{cai2020reconstruct}, we have to train different models for different objects separately. On every test object, the reference images for Gen6D are used as training set for PVNet and RLLG. However, only $\sim$200 reference images are not enough to produce reasonable results so we additionally label the object masks on these reference images and cut the object to randomly paste on backgrounds from COCO~\cite{lin2014coco} to enlarge their training set. For PVNet, we use its 8 corners of the 3D bounding box as keypoints for voting because no model is available.



\textbf{Comparison with baselines}. 
1) Both ObjDesc~\cite{wohlhart2015learning} and ``Ours w/o Ref" select the most similar reference image to estimate the object pose. The results show that our viewpoint selector is able to select more accurate viewpoint than ObjDesc. 
However, only selecting the best reference viewpoint is not enough for predicting accurate poses because the reference images do not cover all possible viewpoints.
2) With further pose refinement, our Gen6D estimator is able to produce better results than the instance-specific methods PVNet and RLLG on average.
The main reason is that for PVNet and RLLG, these reference images are not enough for training a very accurate pose estimator. 
In contrast, Gen6D well adapts into this setting with limited reference images of a novel object. 
Our pose refiner is able to learn generalizable features for accurate pose refinement.



\begin{table}[]
    \centering
    \caption{ADD-0.1d on LINEMOD~\cite{hinterstoisser2012linemod} dataset. ``Training" means what kind of training set is used. ``Synthetic" means the model only uses synthetic data of the given object for training; ``Real" means the model is trained on both the synthetic images and real images of the given model; ``No" means the model is not trained on any data of the test object. ``GT-BBox" means a model uses the ground-truth bounding box or not to produce its performance. ``Refine" means the pose refiner.}
    \resizebox{\textwidth}{!}{
    \begin{tabular}{cccc C{0.08\linewidth} C{0.08\linewidth} C{0.08\linewidth}C{0.08\linewidth}C{0.08\linewidth}C{0.08\linewidth}C{0.08\linewidth}}
    \toprule
    \multirow{2}{*}{Training}
    & \multirow{2}{*}{Name} & GT- & \multirow{2}{*}{Refine} & 
    \multicolumn{6}{c}{Object Name} & \multirow{2}{*}{Avg.}\\
    & & BBox & 
                       &  cat  & duck & bvise &  cam  & driller & lamp &\\
    \midrule
    \multirow{4}{*}{Synthetic}
    & AAE~\cite{sundermeyer2018implicit}
    & \xmark & No   & 17.90 &  4.86 & 20.92 & 30.47 & 23.99 & 60.47 & 26.44 \\
    & Self6D~\cite{wang2020self6d}
    & \xmark & No   & 57.90 & 19.60 & 75.20 & 36.90 & 67.00	& 68.20 & 54.13 \\
    & DPOD~\cite{zakharov2019dpod}   
    & \xmark & No   & 32.36 & 26.12 & 66.76 & 24.22 & 66.60	& 57.26 & 45.55 \\
    & DPOD~\cite{zakharov2019dpod}   
    & \xmark & DPOD~\cite{zakharov2019dpod}  
    & 65.10 & 50.04 & 72.69 & 34.76 & 73.32	& 74.27 & 61.70 \\
    \midrule
    \multirow{4}{*}{Real}
    & PFS~\cite{xiao2019pose}    & \cmark &DeepIM~\cite{li2018deepim}
    & 54.10 & 48.60 & 63.80 & 40.00 & 75.30 & 55.30 & 56.18 \\
    & PVNet~\cite{peng2019pvnet}  
    & \xmark &   No & 79.34 & 52.58 & 99.90 & 86.86 & 96.43 & 99.33 & 85.74 \\
    &PoseCNN~\cite{xiang2017posecnn} 
    & \xmark &DeepIM~\cite{li2018deepim}
    & 82.10 & 77.70 & 97.40 & 93.50 & 95.00 & 96.84 & 95.19 \\
    & DPOD~\cite{zakharov2019dpod}   & \xmark & DPOD~\cite{zakharov2019dpod} 
    & 94.71 & 86.29 & 98.45 & 96.07 & 98.80 & 97.50 & 90.53 \\
    \midrule
    \multirow{4}{*}{Gen}
    & PFS~\cite{xiao2019pose}   
    & \cmark  &   No & 15.40 &  8.20 & 25.10 & 12.10 & 18.60 & 6.50  & 14.32 \\
    & Ours  & \cmark  &   No & 94.11 & 81.31 & 99.52 & 94.31 & 96.33 & 93.38 & 93.16 \\
    & Ours  & \xmark  &   No & 15.97 &  7.89 & 25.48 & 22.06 & 17.24 & 35.80 & 20.74 \\
    & Ours  & \xmark  &Volume& 60.68 & 40.47 & 77.03 & 66.67 & 67.39 & 89.83 & 67.01 \\
    
    \bottomrule
    \end{tabular}
    }
    \label{tab:linemod_add}
\end{table}

\subsection{Results on LINEMOD~\cite{hinterstoisser2012linemod}} 
\begin{figure}
    \centering
    \begin{tabular}{cccc}
    \includegraphics[width=0.22\textwidth]{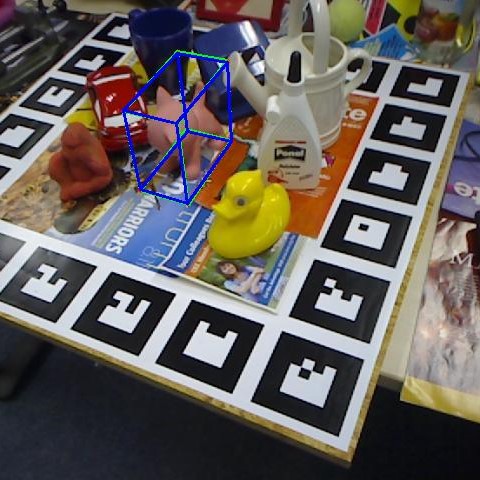} &
    \includegraphics[width=0.22\textwidth]{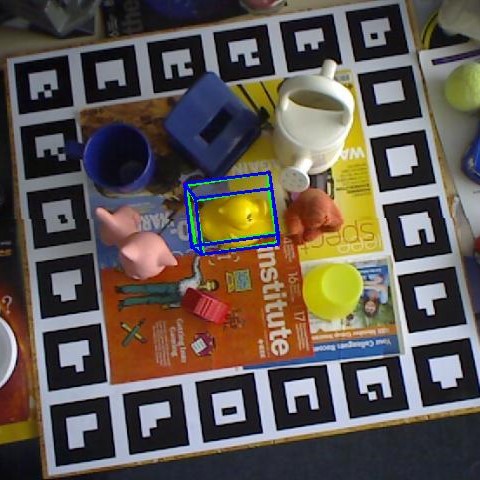} &
    \includegraphics[width=0.22\textwidth]{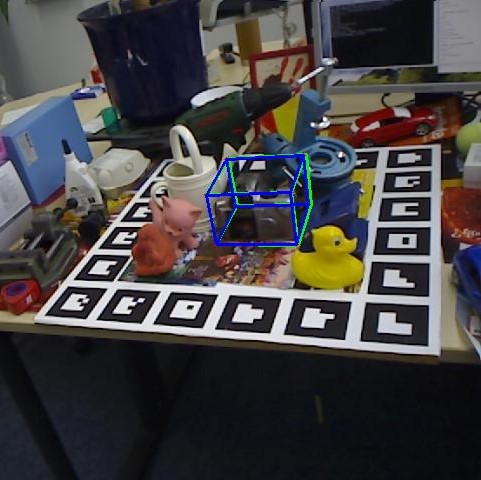} &
    \includegraphics[width=0.22\textwidth]{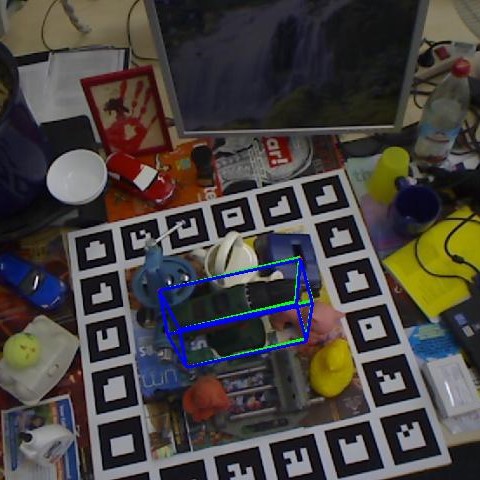} \\
    \end{tabular}
    \caption{Qualitative results on the LINEMOD~\cite{hinterstoisser2012linemod} dataset. Ground-truth poses are drawn in green while predicted poses of Gen6D are drawn in blue.}
    \label{fig:linemod}
\end{figure}

We further report results in ADD-0.1d on the LINEMOD~\cite{hinterstoisser2012linemod} dataset in Table~\ref{tab:linemod_add} and show qualitative results in Fig.~\ref{fig:linemod}. For baselines, we include the instance-specific pose estimators~\cite{sundermeyer2018implicit,wang2020self6d,zakharov2019dpod,cai2020reconstruct,peng2019pvnet,xiang2017posecnn,li2018deepim} and a generalizable estimator Pose-From-Shape (PFS)~\cite{xiao2019pose}. The instance-specific estimators are either trained on the synthetic data of the object (``synthetic training")~\cite{sundermeyer2018implicit,wang2020self6d,zakharov2019dpod} or trained on both the synthetic and real data of the object (``real training")~\cite{peng2019pvnet,xiang2017posecnn,zakharov2019dpod}. PFS~\cite{xiao2019pose} is trained on ShapeNet~\cite{chang2015shapenet}, which embeds an object shape into a feature vector and applies the embedded feature vector on a query image to predict an object pose. For \cite{xiao2019pose,zakharov2019dpod,xiang2017posecnn}, we also include their reported performance using pose refiners DeepIM~\cite{li2018deepim} or DPOD~\cite{zakharov2019dpod}, both of which are trained on the synthetic data or real data of the test object. Ground-truth bounding box is used in PFS to crop the object region for the pose estimation. For baselines, we use the performance reported in their paper for comparison. 

The results in Table~\ref{tab:linemod_add} show that: 1) In comparison with the generalizable pose estimator PFS~\cite{xiao2019pose}, Gen6D outperforms PFS~\cite{xiao2019pose} with or without subsequent pose refinement. Note the PFS~\cite{xiao2019pose} uses the DeepIM~\cite{li2018deepim} refiner which actually is trained on the synthetic data of the test object while our volume-based refiner is not trained on the test object at all. 2) In comparison with instance-specific estimators~\cite{zakharov2019dpod,wang2020self6d,sundermeyer2018implicit} with synthetic training on the test object, Gen6D clearly outperforms all these methods. 3) However, Gen6D performs worse than instance-specific estimators~\cite{peng2019pvnet,xiang2017posecnn,zakharov2019dpod} with real training. The main reason is the inaccurate estimation of the depth. Since the object is usually very far away from the camera and small scale difference (1-2 pixels) will result in a huge offset in the depth direction. Without training on the object, Gen6D cannot perceive such subtle scale changes, which results in worse performance. 4) With ground-truth bounding box, Gen6D achieves comparable results as the instance-specific estimators~\cite{peng2019pvnet,xiang2017posecnn,zakharov2019dpod} with real training because such ground-truth bounding boxes provide correct depths.

\begin{table}[]
    \centering
    \caption{ADD-AUC on the MOPED dataset with threshold 0-10cm. ``LF" means Latent-Fusion~\cite{park2020latentfusion}. ``General" means the pose estimator is trained on the specific object or not. ``Input" means the required type of query images at test time.}
    \begin{tabular}{lcccccccc}
    \toprule
    \multirow{2}{*}{Method} & \multirow{2}{*}{General} & \multirow{2}{*}{Input} & \multicolumn{5}{c}{Object Name} & \multirow{2}{*}{avg.} \\
     & & & B.Drill & D.Dude & V.Mug & T.Plane & R.Aid &  \\
    \midrule
    LF~\cite{park2020latentfusion}  & \cmark & RGBD 
    & \textbf{74.11}  & \textbf{75.40} & 38.27 & 54.95 & 62.97 & 61.14\\
    PVNet~\cite{peng2019pvnet}  & \xmark & RGB 
    & 49.49 & 43.30 & \textbf{67.78} & 48.61 & \textbf{72.92} & 56.42 \\
    Ours & \cmark & RGB 
    & 64.87 & 59.23 & 50.95 & \textbf{69.83} & 72.03 & \textbf{63.38}\\
    \bottomrule
    \end{tabular}
    \label{tab:moped}
\end{table}

\subsection{Results on MOPED~\cite{park2020latentfusion}}

On the MOPED dataset, we compare Gen6D with Latent-Fusion~\cite{park2020latentfusion} and PVNet \cite{peng2019pvnet}. Latent-Fusion~\cite{park2020latentfusion} is also a generalizable pose estimator which does not require training on the test object but needs depth and object masks on query images. We use the official codes and the pretrained weights of Latent-Fusion~\cite{park2020latentfusion} for evaluation. 
For training PVNet~\cite{peng2019pvnet}, we apply the same strategy as used on the GenMOP dataset. 
Table~\ref{tab:moped} reports ADD-AUC on the MOPED dataset, which shows that Gen6D outperforms both baselines on average while Gen6D only uses simple RGB inputs and does not require training on the object.

\subsection{Analysis}

\textbf{Ablation study on the viewpoint selector}. To demonstrate the designs in the viewpoint selector, we conduct ablation studies on the GenMOP dataset and results are shown in Table~\ref{tab:ablation}. Without global normalization and reference view transformer, our viewpoint selector already outperforms the baseline image embedding method ObjDesc~\cite{wohlhart2015learning} by a large margin because our selector pixel-wisely compare the query image with reference images to compute similarity scores, which is more robust to clutter backgrounds. Meanwhile, adding the global normalization or the reference view transformer further improves the results because they exchange information between reference images to help the selector choose the relatively most similar reference image.

\begin{table}[]
    \centering
    \caption{Ablations on the GenMOP dataset. ``GN" means the global normalization used in view selector. ``RVT" means the reference view transformer. ``+ DeepIM Ref." means using the refiner DeepIM~\cite{li2018deepim} to refine the pose for one step. ``+Volume Ref." means refinement with our volume-base refiner for one step.}
    \begin{tabular}{C{0.15\linewidth} lcccccc}
    \toprule
    \multirow{2}{*}{Metrics} & \multirow{2}{*}{Method}  &  \multicolumn{5}{c}{Object Name} & \multirow{2}{*}{avg.}\\
    & & Chair & PlugEN & Piggy & Scissors & TFormer & \\
    \midrule
    \multirow{6}{*}{ADD-0.1d}
    & ObjDesc~\cite{wohlhart2015learning}  
    &   3.50& 5.14& 14.07& 1.25& 7.54& 8.55\\
    & w/o GN and RVT 
    & 8.50 & 13.08 & 36.18 & 14.66 & 1.98 & 14.88\\
    & w/o RVT 
    & 14.50 & 10.75 & 36.18 & 14.22 & 11.51 & 17.43\\
    & Full selector
    & 14.00 & 7.48 & 39.70 & 16.81 & 11.51 & 17.90\\
    &+ DeepIM Ref. & 12.50 & 6.54 & 29.15 & 18.10 & 31.35 & 19.53 \\
    &+ Volume Ref. & 50.50& 9.81& 55.28& 24.57& 52.78& 38.59\\
    \midrule
    \multirow{6}{*}{Prj-5}
    & ObjDesc~\cite{wohlhart2015learning}    
    &  4.00&  10.75&  4.52& 18.53& 8.33& 9.23\\
    & w/o GN and RVT 
    & 7.00 & 40.19 & 20.60 & 28.88 & 54.76 & 30.28\\
    & w/o RVT 
    & 16.00 & 46.73 & 31.66 & 24.57 & 55.16 & 34.82\\
    & Full selector
    & 11.50 & 40.65 & 33.17 & 34.05 & 64.29 & 36.73\\
    &+ DeepIM Ref. & 4.50 & 5.23 & 18.50 & 61.64 & 73.81 & 42.16 \\
    &+ Volume Ref. & 44.00& 71.03& 92.96& 84.48& 95.24& 77.54 \\
    \bottomrule
    \end{tabular}
    \label{tab:ablation}
\end{table}

\textbf{Analysis on the pose refiner}.
To demonstrate the advantage of our volume-based refiner on unseen objects over other rendering-and-comparison based refiners~\cite{li2018deepim,zakharov2019dpod,sundermeyer2020multi}, we report results on the GenMOP in Table~\ref{tab:ablation}.
For the baseline refiner DeepIM~\cite{li2018deepim}, we regard the reference image selected by our selector as the rendered image and use DeepIM to match it with the query image to update the pose, i.e. one step refinement. Note further refinement with more steps using DeepIM are infeasible because there is no object model to render a new image on the updated pose. The DeepIM refiner is trained on the same training data as our volume-based refiner. The results show that our volume-based refiner has better generalization ability on unseen objects than DeepIM. 

\textbf{More analysis}. We provide more analysis about reference image number, comparison with finetuned DeepIM~\cite{li2018deepim}, refinement iterations, ablation on training data and symmetric objects in the supplementary material.

\textbf{Limitations}. The generalization ability of Gen6D mainly comes from matching image patterns in the viewpoint selection and the pose refinement. Thus, Gen6D requires enough diverse training data to learn general image matching for accurate pose estimation and it performs worse with limited training data. Meanwhile, Gen6D is not specially designed to handle occlusions and the performance may degenerate when severe occlusions exist.


\textbf{Running time}. To process an image of size 540$\times$960, Gen6D estimator costs $\sim$0.64 second in total on a 2080Ti GPU, in which the object detector costs $\sim$0.1 second, the viewpoint selector costs $\sim$0.04 second and the refiner with 3 times refinement costs $\sim$0.5 second.

\section{Conclusion}
In this paper, we propose an easy-to-use 6-DoF pose estimator Gen6D for unseen objects. To predict poses for unseen objects, Gen6D does not require the object model but only needs some posed images of the object to predict its pose in arbitrary environments. 
In Gen6D, we design a novel viewpoint selector and a novel volume-based pose refiner. 
Experiments demonstrate the superior performance of Gen6D estimator in predicting poses for unseen objects in the model-free setting.

\clearpage

\bibliographystyle{splncs04}
\bibliography{egbib}

\newpage
\appendix
\vspace{5pt}
\section{Overview} 
This supplementary material has the following contents.
\begin{enumerate}
    \item Sec.~\ref{sec:detail} contains implementation details of the data normalization (Sec.~\ref{sec:detail_norm}), the object detector (Sec.~\ref{sec:detail_det}), the viewpoint selector (Sec.~\ref{sec:detail_sel}), the pose refiner (Sec.~\ref{sec:detail_ref}) and the inference process.
    \item Sec.~\ref{sec:prot} includes experiment details about the training set, the GenMOP dataset and the reference/query splits.
    \item Sec.~\ref{sec:quality} provides more qualitative results on the LINEMOD~\cite{hinterstoisser2012linemod} dataset, the MOPED~\cite{park2020latentfusion} dataset and the GenMOP dataset.
    \item Sec.~\ref{sec:more_analysis} contains additional analysis on comparison with finetuned DeepIM~\cite{li2018deepim}, reference image number, unevenly distributed reference images, refinement iterations, ablations on training data, symmetric objects and imperfect view selection.
\end{enumerate}

\section{Implementation detail}
\label{sec:detail}
\subsection{Data normalization}
\label{sec:detail_norm}

\textbf{Object size and center}. On the GenMOP dataset, the object size and object center are roughly estimated from reconstructed points of COLMAP~\cite{schoenberger2016colmap}. On the LINEMOD~\cite{hinterstoisser2012linemod} dataset and the MOPED~\cite{park2020latentfusion} dataset, we directly use the provided 3D model to compute the object size and the object center.
Note we do not need an exact object size and an object center.

\textbf{Normalization of object coordinate system}.
Given the object size $d$ and the object center $c$, we normalize the object coordinate system by
\begin{equation}
    x_{norm}=(x-c)/d \times 2,
\end{equation}
where $x_{norm}$ is the normalized coordinate and $x$ is the original coordinate.

\begin{figure}
    \centering
    \includegraphics[width=0.6\textwidth]{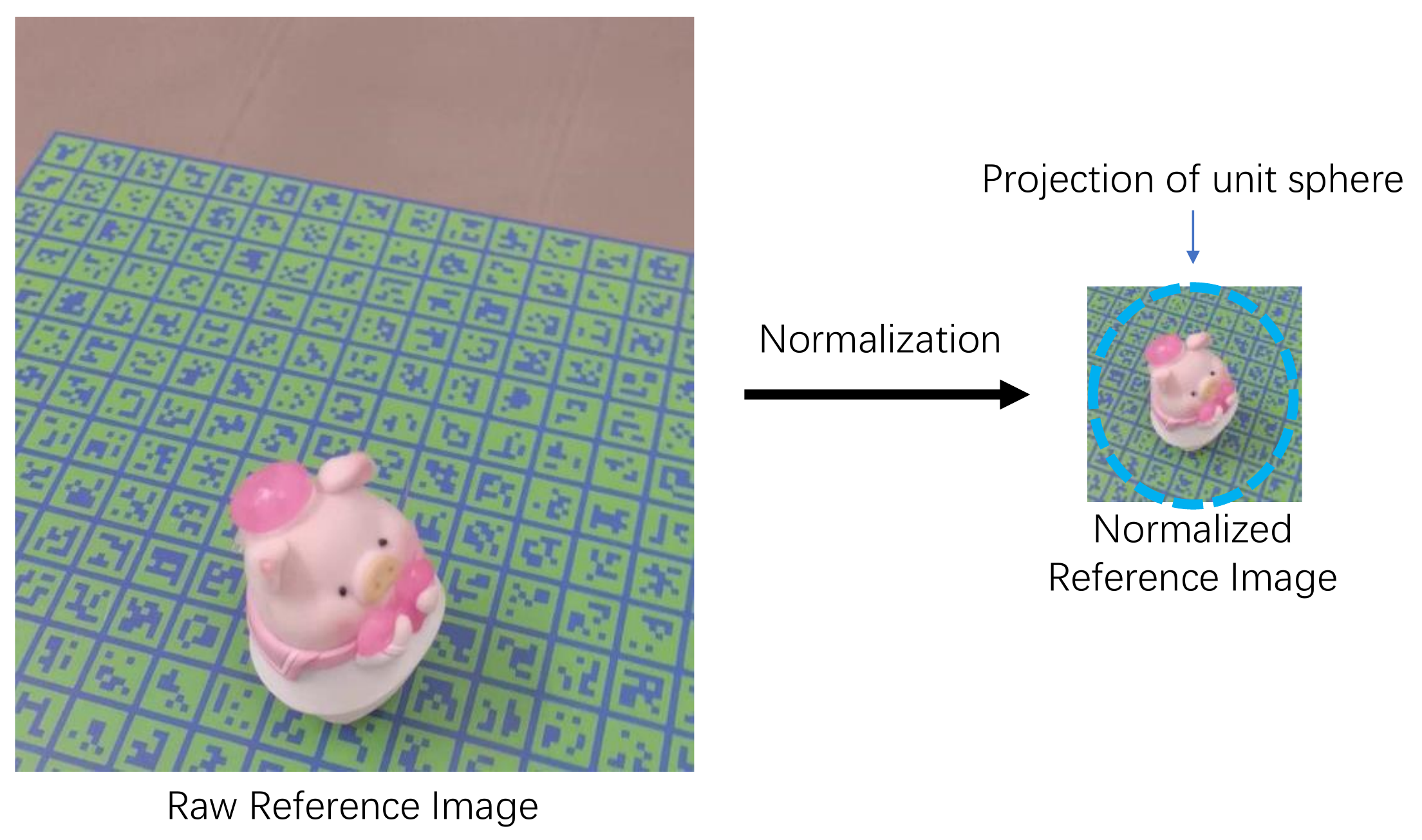}
    \caption{Raw reference images are normalized to look at the object. The projections of the unit sphere are enclosed by the reference images.}
    \label{fig:ref_norm}
\end{figure}

\textbf{Normalization of reference images}.
Since the raw reference images may not look at the object, we normalize these reference images by warping the image with a homography and changing the intrinsic correspondingly, as shown in Fig.~\ref{fig:ref_norm}. The normalized reference images all look at the object and they enclose the projection of the unit sphere at the origin. Note we can do this normalization because we have already normalized the object coordinate and we know the poses of reference images.

\subsection{Detector}
\label{sec:detail_det}
\begin{figure}
    \centering
    \includegraphics[width=\textwidth]{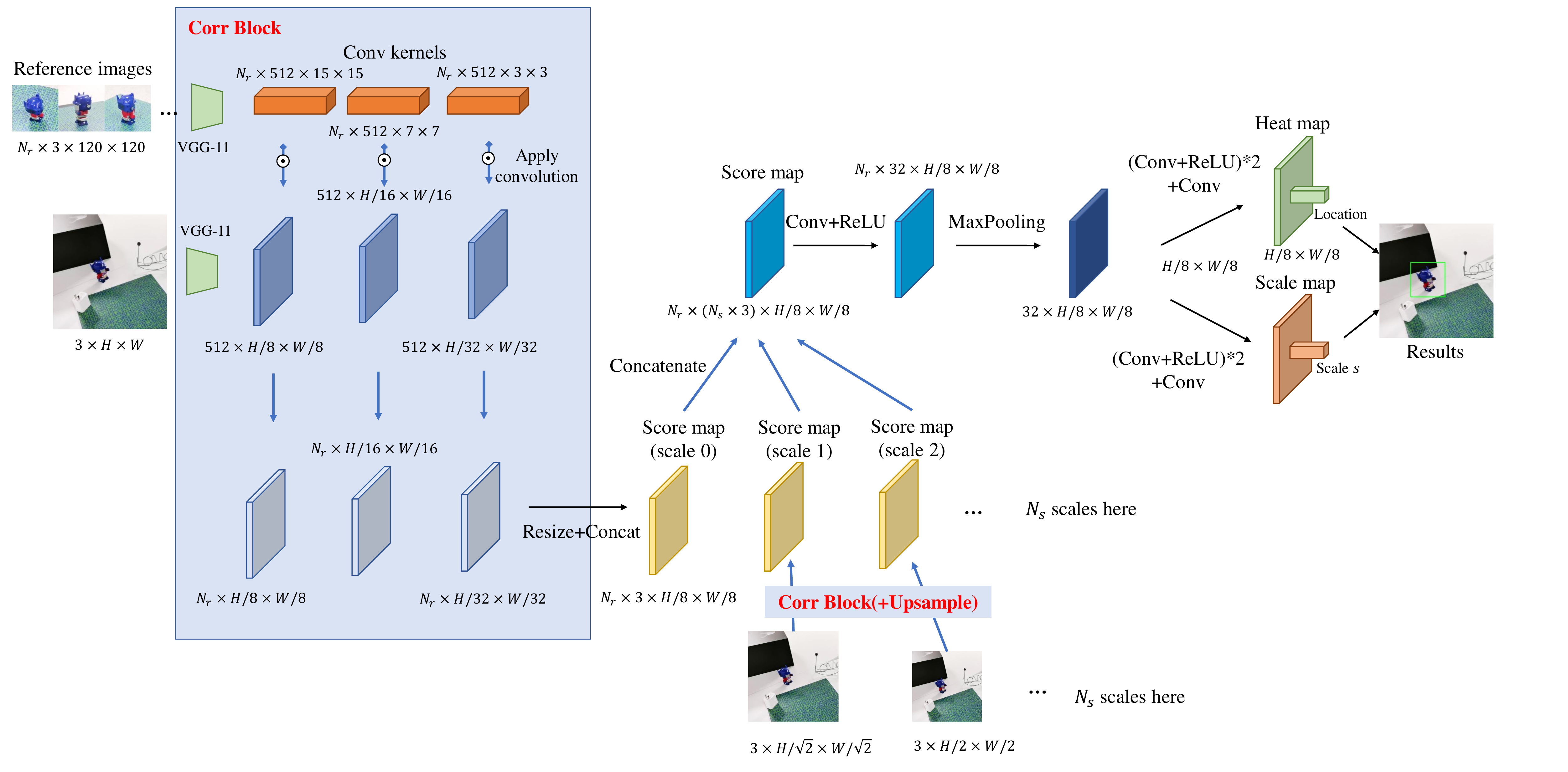}
    \caption{Detailed pipeline of the detector. All ``Conv" layers use 3$\times$3 kernel size.}
    \label{fig:det_pipeline}
\end{figure}

\textbf{Networks architecture}.
The detailed architecture is shown in Fig.~\ref{fig:det_pipeline}. The input reference images are resized to $120\times120$. We actually use three levels of feature maps from the query image and the corresponding three levels of Conv kernels from reference images. We conduct convolutions on the feature map at each level with the corresponding conv kernel at the same level.
$N_s=5$ scales are used and we downsample the input query image with $\sqrt{2}$ as the factor.

\textbf{Loss}. Given the heat map $H$, we regard all values of the heat map as logits and use a binary classification loss to supervise the heat map prediction. Here, we first project the object center to the image using the ground-truth object pose. Then, a pixel on the heat map is assumed to be correct if it is within 1.5-pixel distance from the object center projection. Otherwise, it is incorrect.
\begin{equation}
    \ell_{heat}=\sum_p -\mathbbm{1}(\|p-c_{prj}\|_2<1.5)\log \sigma(H(p))-(1-\mathbbm{1}(\|p-c_{prj}\|_2<1.5))\log(1-\sigma(H(p))),
\end{equation}
where $\mathbbm{1}$ is an indicator function, $c_{prj}\in\mathbb{R}^2$ is the 2D projection of the object center, $p$ is a pixel on the heat map, $\sigma$ is the Sigmoid function, $H(p)$ means the heat value on the pixel $p$.
To supervise the scale map prediction, we apply a scale loss to minimize the L2 distance between the ground-truth scale and the predicted scale in the log space. 
Note we only apply such scale loss on the pixels within 1.5-pixel distance away from the projected object center.
\begin{equation}
    \ell_{scale}=\sum_p \mathbbm{1}(\|p-c_{2d}\|_2<1.5) \|\log{s_{gt}}- S(p)\|_2^2,
\end{equation}
where $s_{gt}$ is the ground-truth scale, $S$ is the scale map, $S(p)$ means the scale value at pixel $p$. $s_{gt}$ can be computed from the distance $l$ between the camera center and the object center by
\begin{equation}
    s_{gt} = \frac{2f}{lS_r},
\end{equation}
where $f$ is a virtual focal length by changing the principle point to the object center projection $c_{prj}$, 2 is the diameter of the unit sphere and $S_r=120$ is the size of reference image. 

\subsection{Selector}
\label{sec:detail_sel}

\begin{figure}
    \centering
    \includegraphics[width=\textwidth]{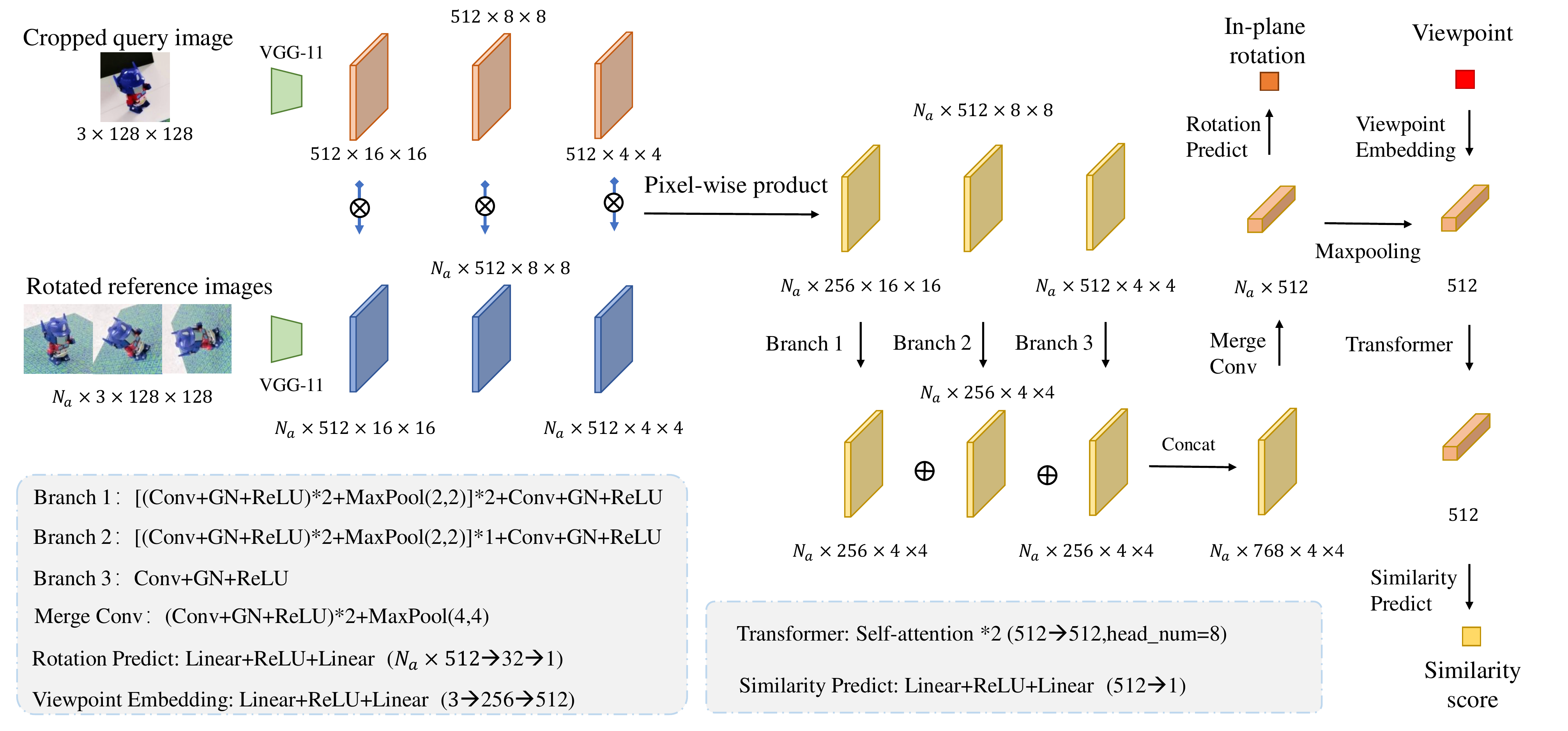}
    \caption{Detailed pipeline of the selector. We only draw one reference image in the figure for clear visualization. ``Conv" means a convolution layer with 3$\times$3 kernels. ``GN" means the global normalization that normalize the feature maps using the mean and variance computed from feature maps of all reference images. ``Transformer" is applied among all feature vectors of reference images.}
    \label{fig:sel_pipeline}
\end{figure}

\textbf{Network architecture}. The detailed architecture of the selector is shown in Fig.~\ref{fig:sel_pipeline}. The input query image is cropped according to the detection results and resized to $128\times128$. All reference images are also cropped to the size 128$\times$128. We also use three levels of feature maps to conduct the pixel-wise product. The viewpoint is represented by a 3-dimensional vector. The in-plane rotation is a scalar of the clock-wise rotation angle. The final similarity score for a reference image is also a scalar. We use $N_a=5$ rotation angles in $[-\pi/2,\pi/2]$ to rotate every reference image.

\textbf{Loss}. To compute the loss, we first introduce how we compute the ground-truth similarity between two viewpoints. In the object coordinate system, we denote the camera locations of all reference images as $\{u_{i}\in \mathbb{R}^3|i=1,...,N_r\}$ where $N_r$ is the number of reference images. Note that the object center is the origin so that $-u_i$ is the vector from the camera to the object center. In training, we know the query camera pose and compute the camera location $v$ of the query image in the object coordinate system. Both $u_i$ and $v$ are normalized by $\tilde{u}_i=u_i/\|u_i\|_2$ and $\tilde{v}=v/\|v\|_2$. Then, the ground-truth viewpoint similarity between i-th reference image and the query image is their dot product $\tilde{u}_i\cdot\tilde{v}$.

We normalize the ground-truth viewpoint similarity to [0,1], which is denoted as $\tilde{s}_j$. Then, the similarity loss is
\begin{equation}
    \ell_{sim}=\sum_j BCE(s_j,\tilde{s}_j)
\end{equation}
where BCE means binary cross entropy loss and we force the predicted score to be consistent with the ground-truth viewpoint similarity.

To train the in-plane rotation angle prediction, we adopt a L2 loss,
\begin{equation}
    \ell_{angle} = \|\alpha_j-\alpha_{gt}\|_2^2,
\end{equation}
where $\alpha_j$ is the predicted in-plane angle on the ground-truth nearest reference image and $\alpha_{gt}$ is the ground-truth in-plane rotation between the query image and the reference image.

\subsection{Refiner}
\label{sec:detail_ref}
\begin{figure}
    \centering
    \includegraphics[width=\textwidth]{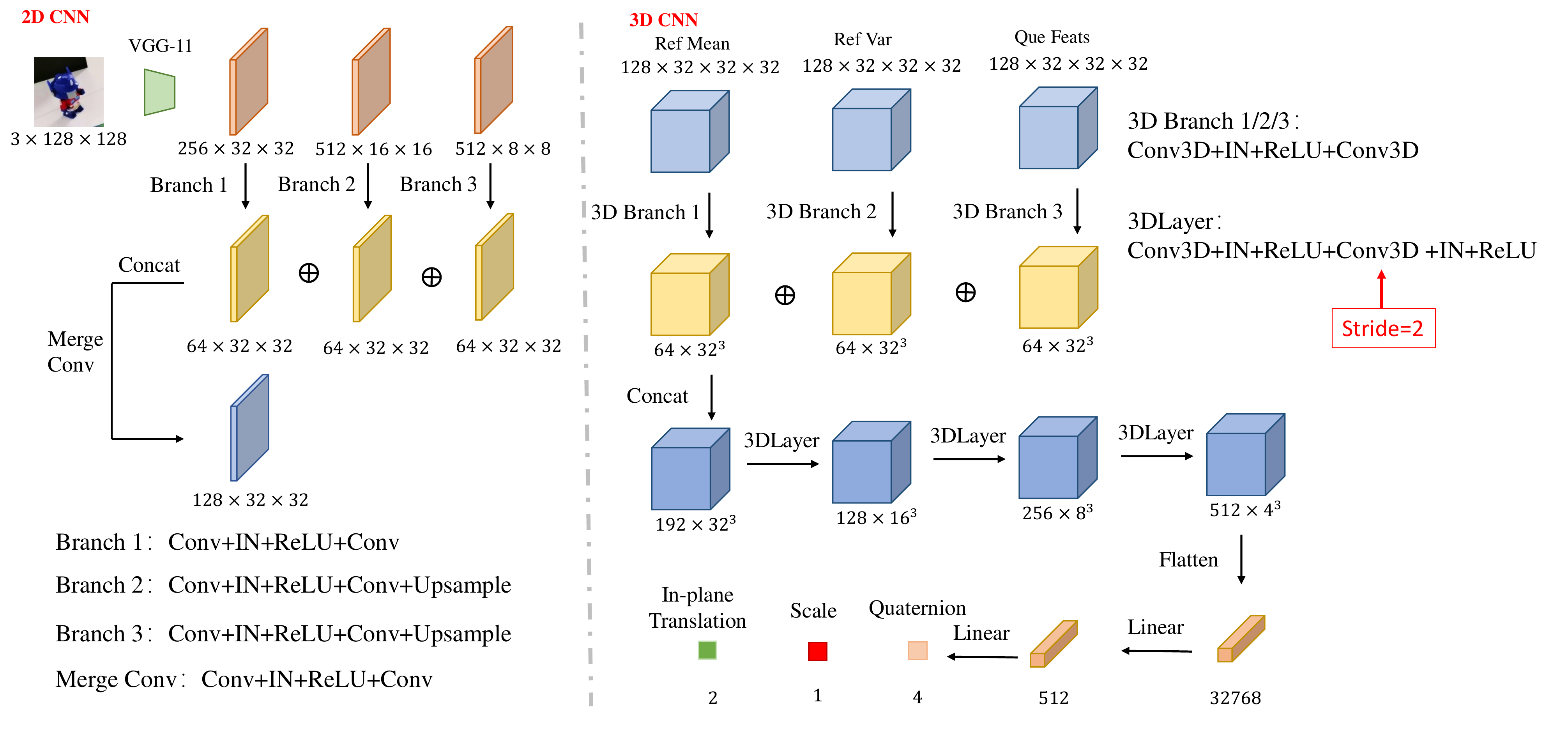}
    \caption{Detailed architecture of the refiner. ``IN" means instance normalization.}
    \label{fig:ref_pipeline}
\end{figure}

\textbf{Network architecture}.
The detailed architecture for the refiner is shown in Fig.~\ref{fig:ref_pipeline}. In the figure, we show the 2D CNN and the 3D CNN separately. In the 2D CNN, both reference images and query images are resized to 128$\times$128 and the final feature map for unprojection is extracted from three levels of feature maps. In the 3D CNN, we first embed reference mean, reference variance and query features separately. Then, they are concatenated and processed by several 3D convolution layers. Final 3D pose residuals are regressed by linear layers from the flatten feature vector of the 3D volume. We use $N_n=6$ neighboring reference images in refinement.

\begin{wrapfigure}{r}{0.5\textwidth}
    \centering
    \includegraphics[width=0.5\textwidth]{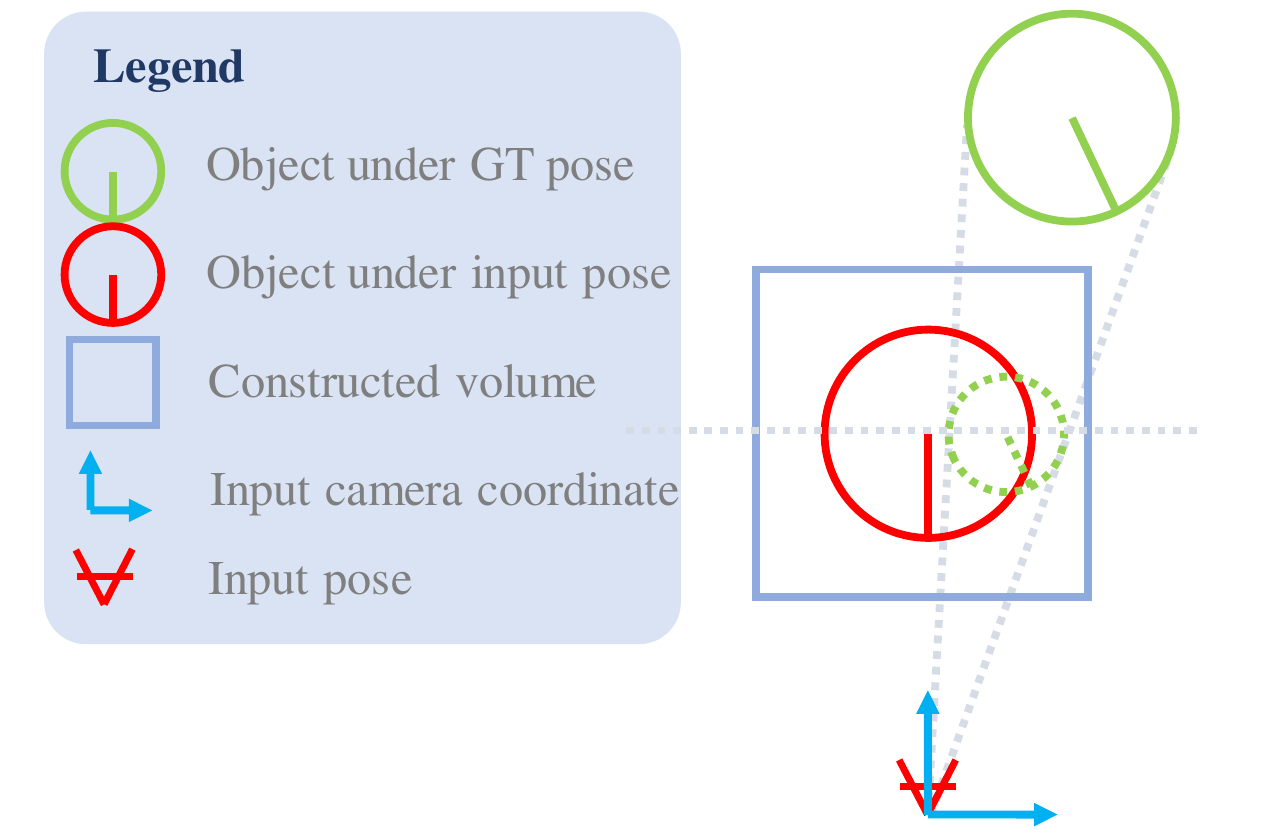}
    \caption{Diagram to illustrate the transformation. Note all transformations are applied on coordinates in the input camera coordinate system.}
    \label{fig:ref_supp}
\end{wrapfigure}

\textbf{Loss}. To train the refiner, we first sample $32^3$ voxel points in the unit cube in the object coordinate system. Then, these points are transformed to the input camera coordinate system by the input pose. The loss is the distance between the sample points transformed by the ground-truth similarity transformation and the sample points transformed by the predicted similarity transformation. 
\begin{equation}
    \ell_{ref}= \sum_k^{32^3} \|s_{pr} R_{pr}  (p_k + t'_{pr}) - s_{gt} R_{gt} (p_k +t'_{gt}) \|_2,
\end{equation}
where $p_k\in \mathbb{R}^3$ is the coordinate of a sample point in the input camera coordinate, $s_{pr}$ and $s_{gt}$ are predicted scale and ground-truth scale respectively, $R_{pr}$ and $R_{gt}$ are the predicted rotation and the ground-truth rotation respectively, $t'_{pr}$ and $t'_{gt}$ are the predicted 2D translation and the ground-with 2D translation respectively. We set the third element of $t'_{pr}$ and $t'_{gt}$ to 0 to form a 3D-translation. Note the predicted and ground-truth similarity transformations are used in transforming the object (red circle in Fig.~\ref{fig:ref_supp}) with the input pose to the object with the ground-truth pose (green dotted circle in Fig.~\ref{fig:ref_supp}).

\textbf{Similarity approximation}. We discuss how to convert the similarity transformation (red circle to dotted green circle in Fig.~\ref{fig:ref_supp}) to the rigid transformation (red circle to solid green circle in Fig.~\ref{fig:ref_supp}). Denoting the similarity transformation as $(R,s,t)$, our target is to compute the rigid transformation $(R,t')$. Actually, we only need to convert $(s,t)$ to $t'$ because rotation $R$ is the same in two transformations. Assuming the red circle center is $(c_x,c_y,c_z)$ in the input camera coordinate, then the center of the green dotted circle is $(c_x+t_x,c_y+t_y,z)$ where $t_x$ and $t_y$ are the first and the second element of $t$ respectively. The depth affects the scale, so the center of the green solid circle is $((c_x+t_x)/s,(c_y+t_y)/s,z/s)$. The final rigid translation is $t'=((c_x+tx)/s-c_x,(c_y+t_y)/s-c_y,z/s-z)$.

\subsection{Inference details}
In inference, we apply pose refiner iteratively for 3 times. For the input data, not all reference images are used in the object detector and the viewpoint selector. Instead, we use farthest point sampling to sample 32 and 64 images for the detector and the selector respectively.

\section{Experimental setting}
\label{sec:prot}
\textbf{Training set for Gen6D}. The training objects on the LINEMOD~\cite{hinterstoisser2012linemod} dataset are ``ape", ``can", ``holepuncher", ``iron" and ``phone". There are $\sim$1200 images per object and we randomly use them as reference images and query images. The training objects on the GenMOP dataset are ``cup", ``knife", ``love", ``plugCN" and ``Miffey", each of which contains $\sim$200 reference images and $\sim$200 query images. 
On every object from ShapeNet~\cite{chang2015shapenet}, we render 1024 images.
On every object from Google Scanned Objects, we use 512 rendered images from IBRNet~\cite{wang2021ibrnet} for training. To train the 2D object detector, we additionally use the CO3D~\cite{reizenstein2021common} for training.

\textbf{Reference/query split on training sets}. On the LINEMOD~\cite{hinterstoisser2012linemod} dataset, the ShapeNet dataset and the Google Scanned objects dataset, we randomly select 128 images by farthest point sampling on camera locations as reference images while the other images are selected as query images. On the GenMOP dataset, we use images from one video as reference images while images from the other video as query images.

\textbf{Reference/query split on test sets}. On the LINEMOD~\cite{hinterstoisser2012linemod} dataset, we use the training set of previous instance-specific estimators~\cite{tekin2018real,peng2019pvnet} as reference images and the other images are selected as query images. On the GenMOP dataset, we also use images from one video sequence as reference images and images from the other video sequence as query images On the MOPED dataset~\cite{park2020latentfusion}, we use the provided reference video sequences as reference images and the other video sequences as query images. We list the number of reference images and query images in Table~\ref{tab:img_num}. 
Note that reference images are used in inference of Gen6D but not in training the Gen6D estimator while instance-specific estimators like PVNet~\cite{peng2019pvnet} actually use these reference images to train their models.

\begin{table}[]
    \centering
    \caption{Numbers of reference images and test query images on different datasets.}
    \begin{tabular}{cccccc}
    \toprule
    & \multicolumn{5}{c}{GenMOP} \\
    & Chair & PlugEN & Piggy & Scissors & TFormer\\
    \midrule
    Reference & 212 & 199 & 227 & 202 & 206 \\
    Query     & 200 & 214 & 199 & 232 & 252 \\
    \midrule
    & \multicolumn{5}{c}{LINEMOD~\cite{hinterstoisser2012linemod}} \\
    & cat  & duck & bvise &  cam  & driller\\
    \midrule
    Reference & 177 & 189 & 183 & 182 & 179 \\
    Query     & 1002 & 1065 & 1032 & 1020 & 1009 \\
    \midrule
    & \multicolumn{5}{c}{MOPED~\cite{park2020latentfusion}} \\
    & B.Drill & D.Dude & V.Mug & T.Plane & R.Aid \\
    \midrule
    Reference & 355 & 451 & 606 & 662 & 394 \\
    Query     & 215 & 297 & 91 & 250 & 58 \\
    \bottomrule
    \end{tabular}
    \label{tab:img_num}
\end{table}

\textbf{GenMOP}. On every object of the GenMOP dataset, we collect two 1-minute videos in different environments by a cell-phone. On each video, we sample 1 image per 10 frames, which results in $\sim$200 images for every video. On each video sequence, we apply COLMAP~\cite{schoenberger2016colmap} to recover the extrinsics and intrinsics of all cameras. Then, in every sequence, we manually select two images, label 4 keypoints on the selected images, and use triangulation to compute the 3D points. For two difference sequences of the same object, we compute the transformation from the triangulated 3D points to align them.

\section{Qualitative Results}
\label{sec:quality}
\begin{figure}
    \centering
    \begin{tabular}{cccc}
    \includegraphics[width=0.22\textwidth]{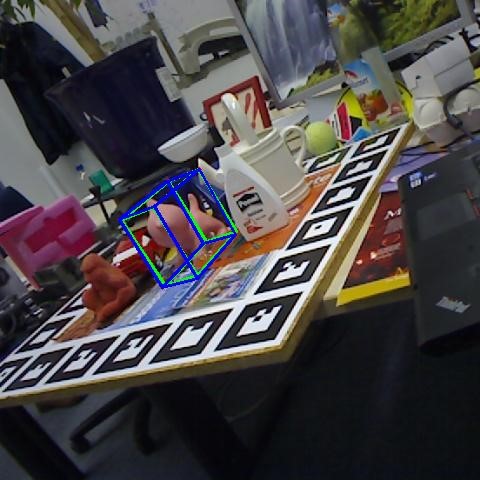} &
    \includegraphics[width=0.22\textwidth]{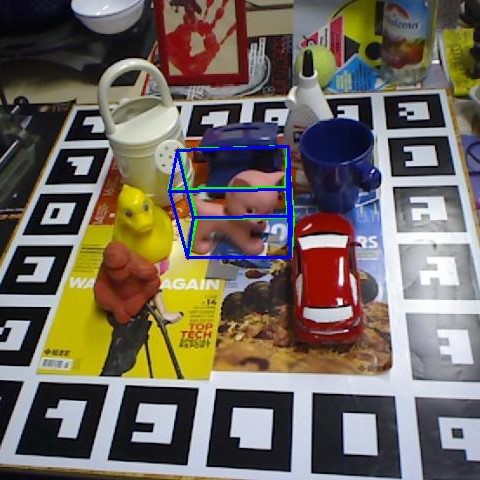} &
    \includegraphics[width=0.22\textwidth]{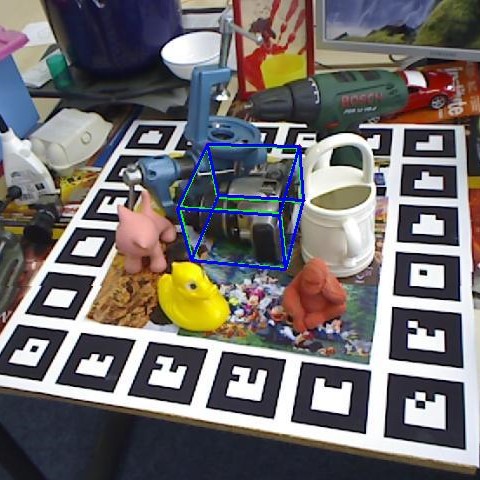} &
    \includegraphics[width=0.22\textwidth]{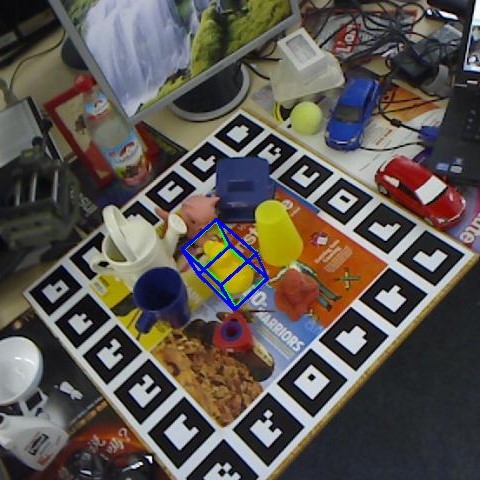} \\
    \includegraphics[width=0.22\textwidth]{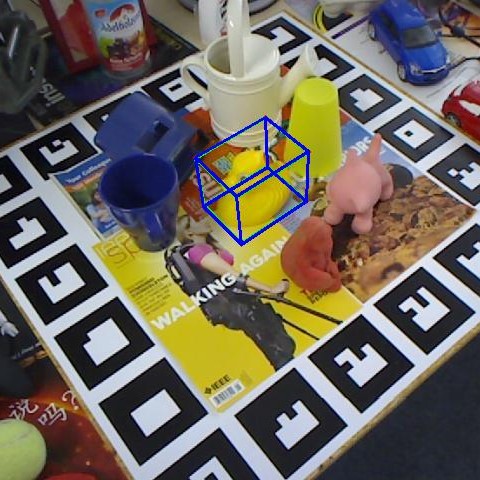} &
    \includegraphics[width=0.22\textwidth]{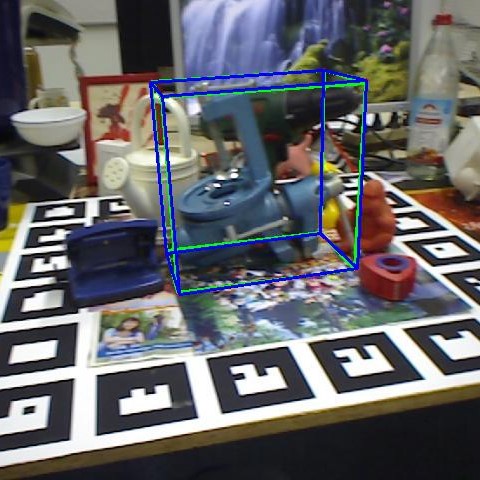} &
    \includegraphics[width=0.22\textwidth]{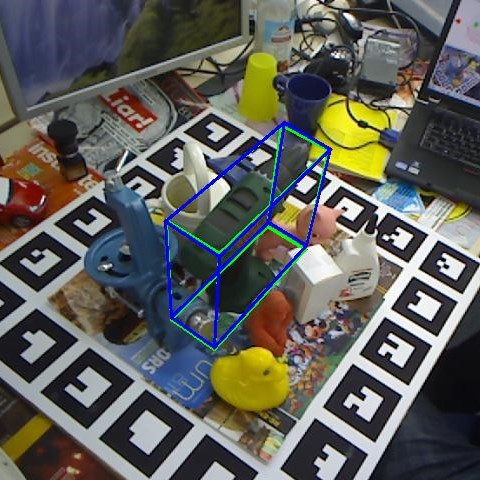} &
    \includegraphics[width=0.22\textwidth]{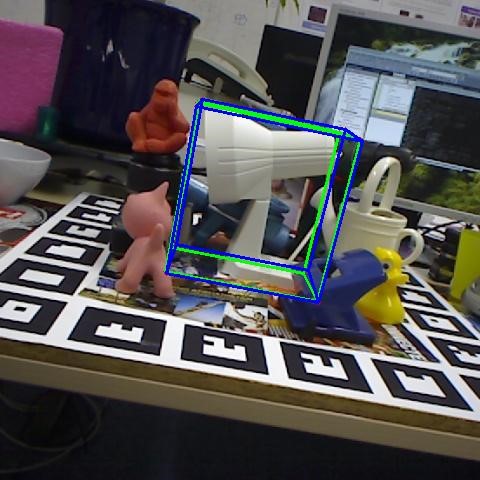} \\
    \end{tabular}
    \caption{Additional qualitative results on the LINEMOD~\cite{hinterstoisser2012linemod} dataset. Ground-truth poses are drawn in green while predicted poses are drawn in blue.}
    \label{fig:linemod_more}
\end{figure}

\begin{figure}
    \centering
    \begin{tabular}{cccc}
    \includegraphics[width=0.22\textwidth]{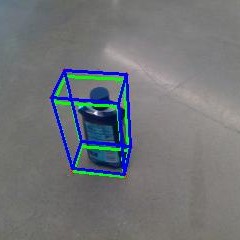} &
    \includegraphics[width=0.22\textwidth]{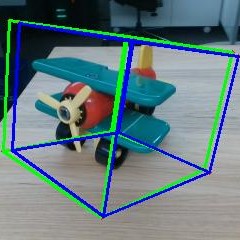} &
    \includegraphics[width=0.22\textwidth]{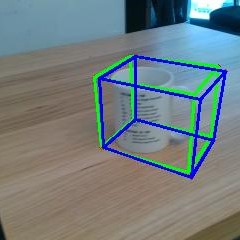} &
    \includegraphics[width=0.22\textwidth]{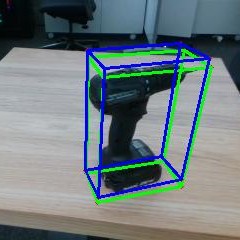}
    \end{tabular}
    \caption{Qualitative results on the MOPED~\cite{park2020latentfusion} dataset. Ground-truth poses are drawn in green while predicted poses are drawn in blue. 
    }
    \label{fig:moped}
\end{figure}

\begin{figure}
    \centering
    \begin{tabular}{cccc}
    \includegraphics[width=0.22\textwidth]{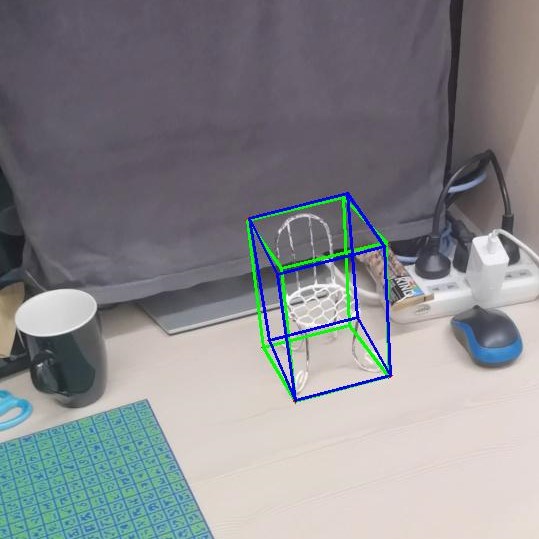} &
    \includegraphics[width=0.22\textwidth]{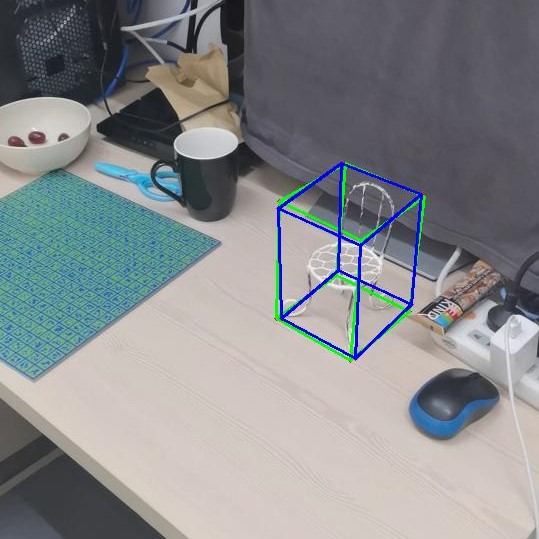} &
    \includegraphics[width=0.22\textwidth]{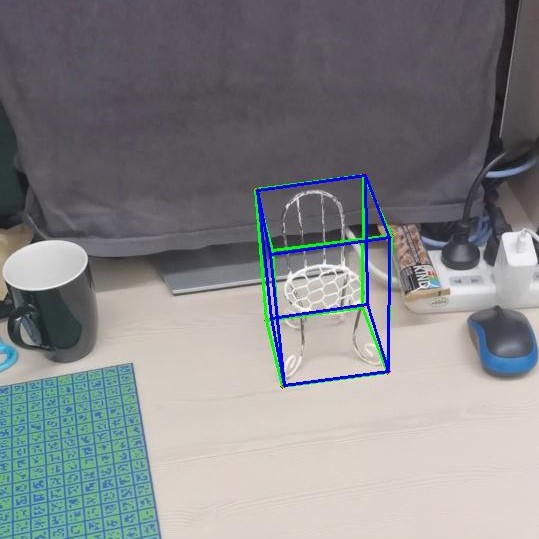} &
    \includegraphics[width=0.22\textwidth]{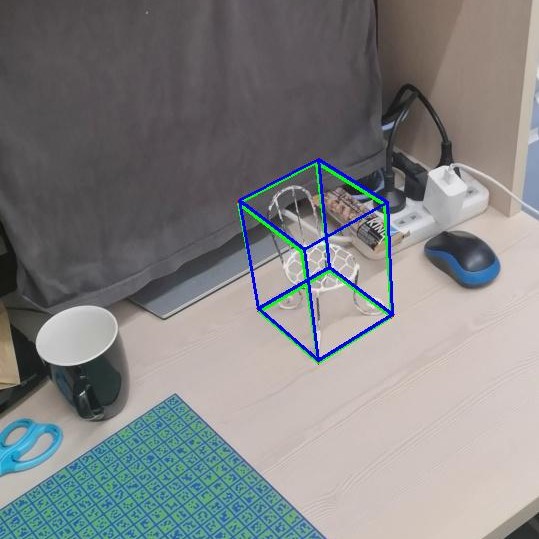} \\
    \includegraphics[width=0.22\textwidth]{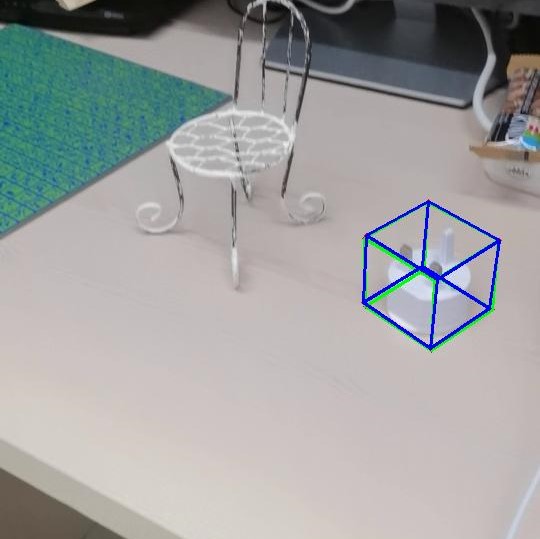} &
    \includegraphics[width=0.22\textwidth]{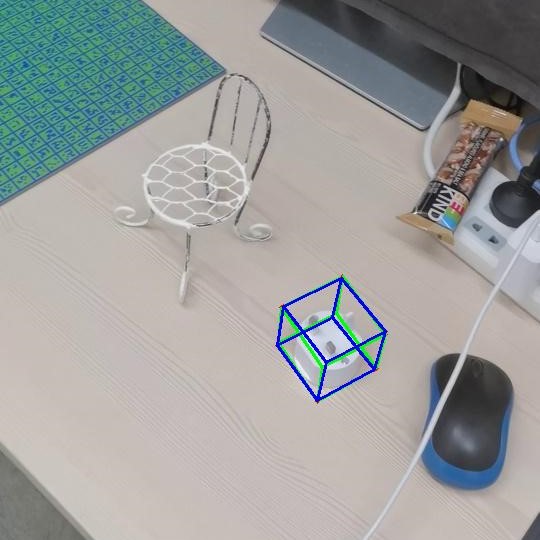} &
    \includegraphics[width=0.22\textwidth]{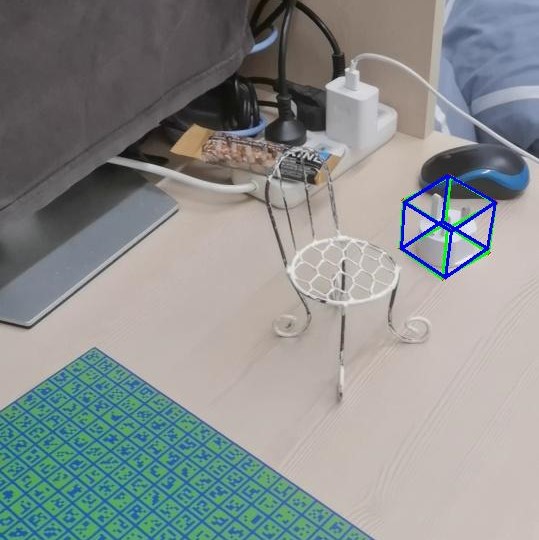} &
    \includegraphics[width=0.22\textwidth]{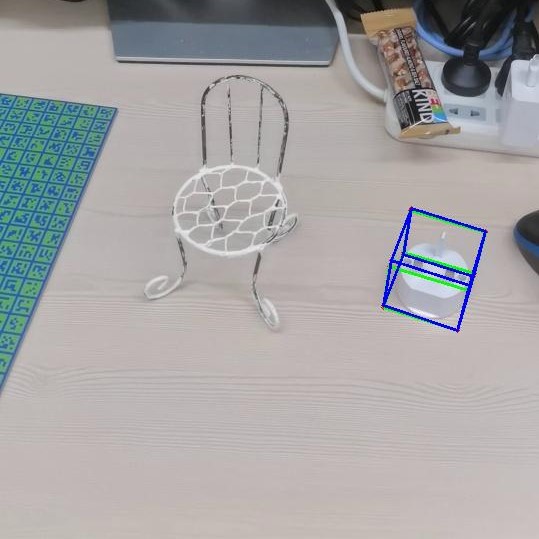} \\
    \includegraphics[width=0.22\textwidth]{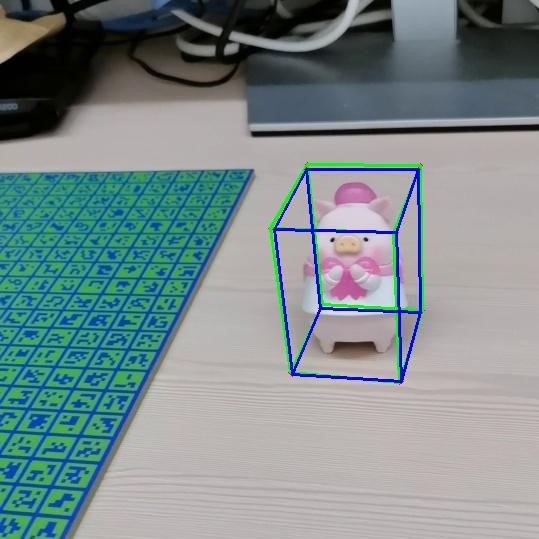} &
    \includegraphics[width=0.22\textwidth]{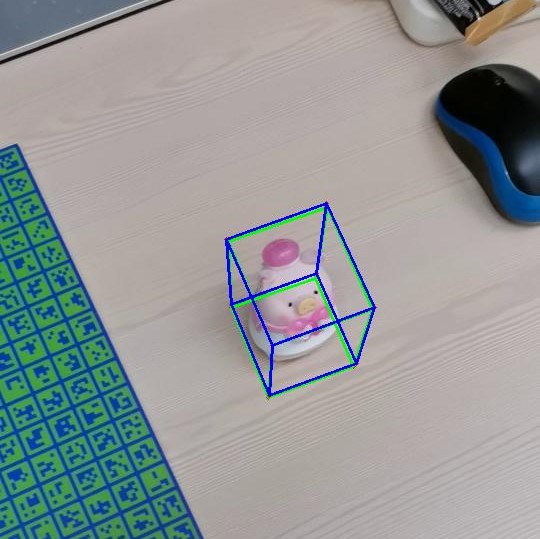} &
    \includegraphics[width=0.22\textwidth]{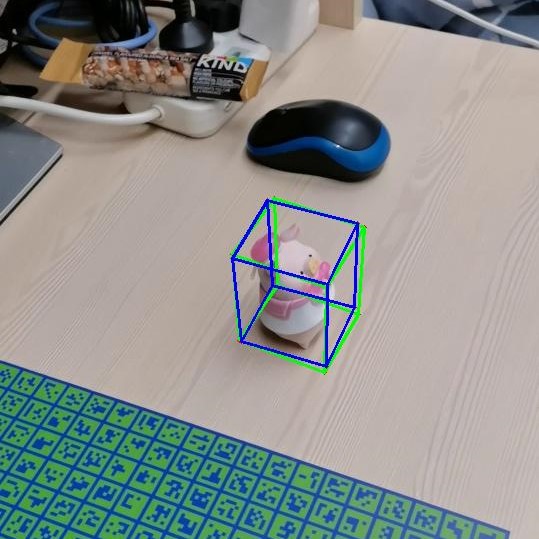} &
    \includegraphics[width=0.22\textwidth]{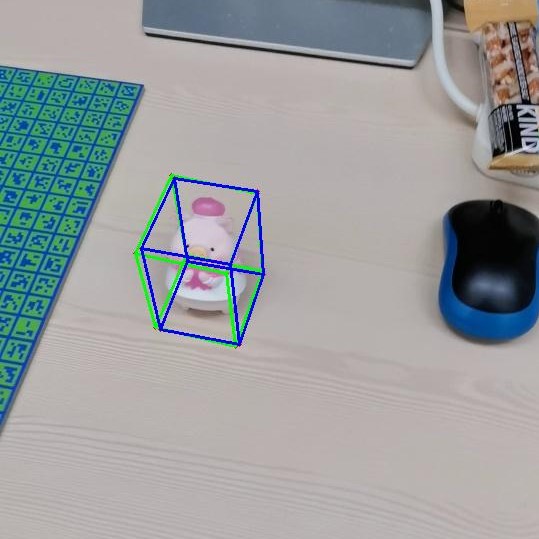} \\
    \includegraphics[width=0.22\textwidth]{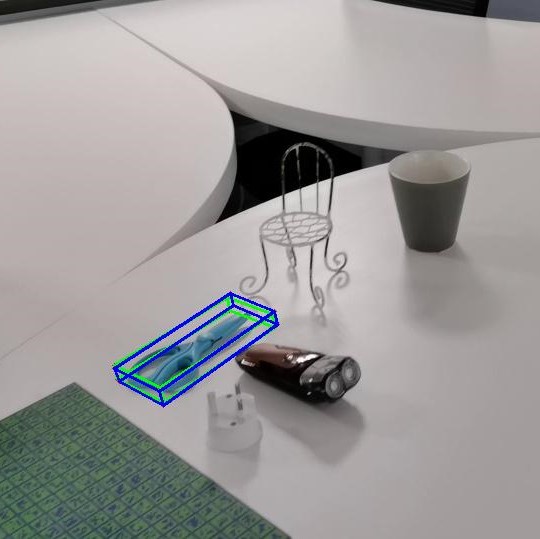} &
    \includegraphics[width=0.22\textwidth]{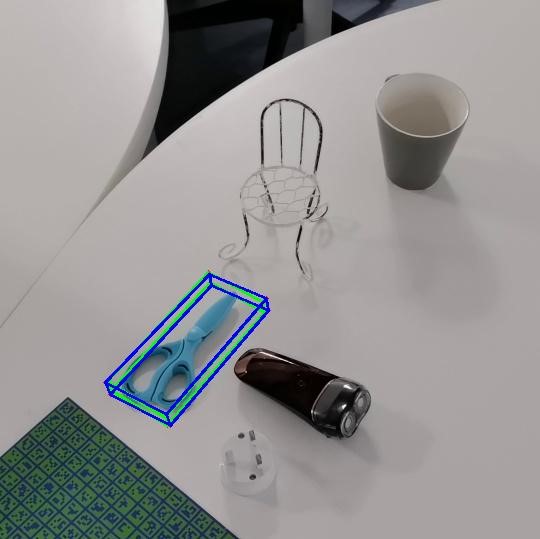} &
    \includegraphics[width=0.22\textwidth]{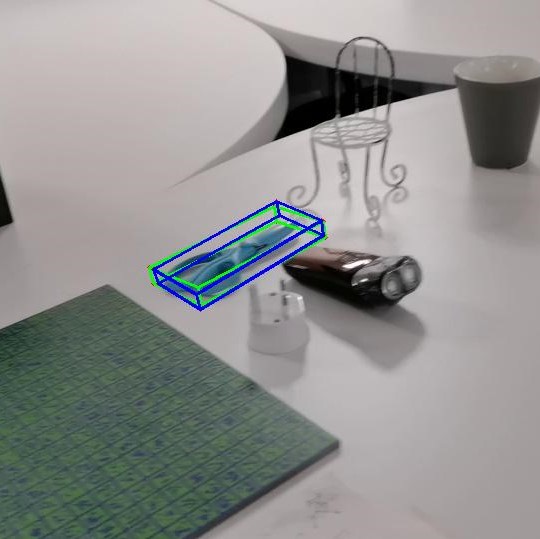} &
    \includegraphics[width=0.22\textwidth]{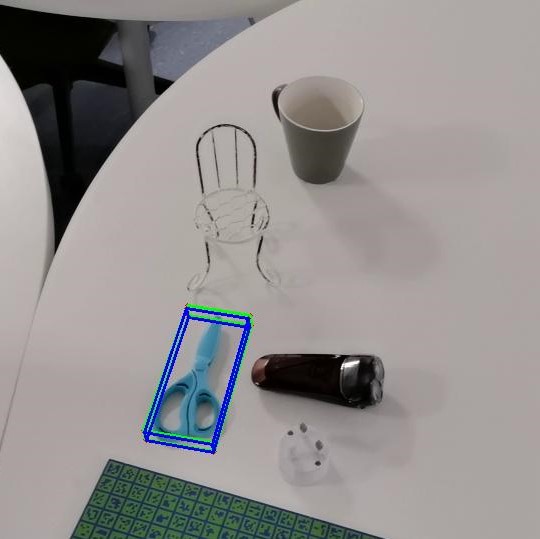} \\
    \includegraphics[width=0.22\textwidth]{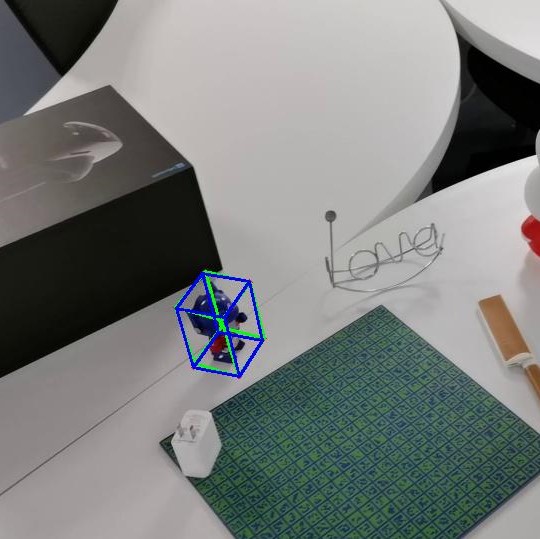} &
    \includegraphics[width=0.22\textwidth]{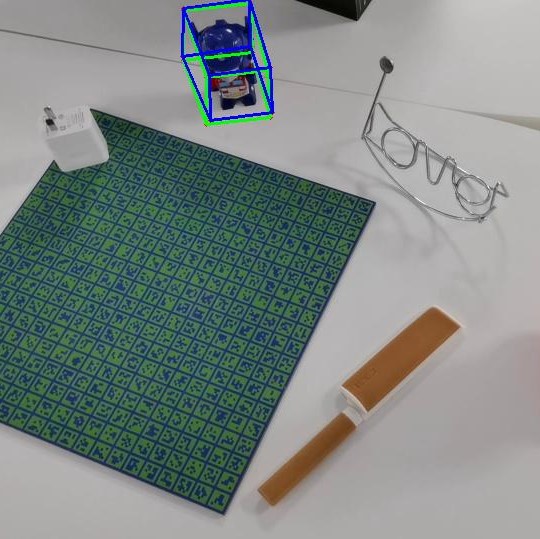} &
    \includegraphics[width=0.22\textwidth]{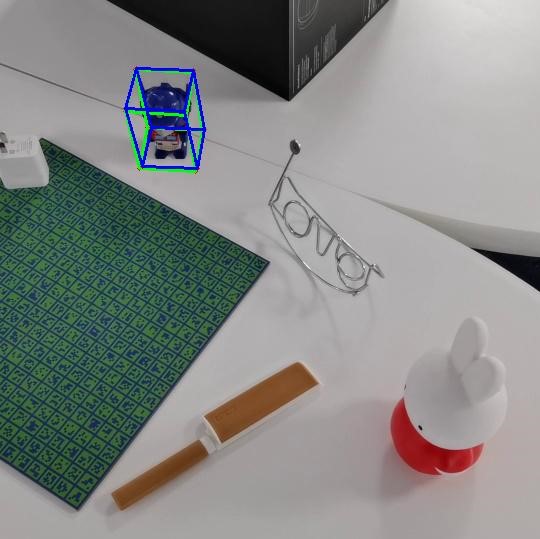} &
    \includegraphics[width=0.22\textwidth]{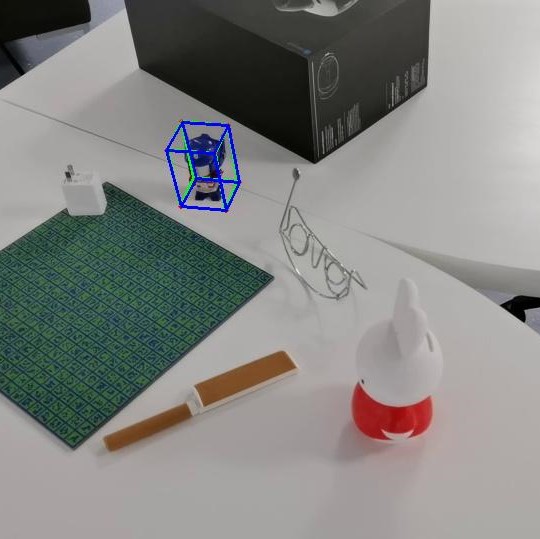} \\
    \end{tabular}
    \caption{More qualitative results of Gen6D on the GenMOP dataset.}
    \label{fig:genmop_more}
\end{figure}

More qualitative results on the LINEMOD dataset, the MOPED~\cite{park2020latentfusion} dataset and the GenMOP dataset are shown in Fig.~\ref{fig:linemod_more}, Fig.~\ref{fig:moped} and Fig.~\ref{fig:genmop_more} respectively.

\section{More analysis}
\label{sec:more_analysis}

\subsection{Comparison with finetuned DeepIM}

Based on the pretrained generalizable DeepIM~\cite{li2018deepim} model in the experiments of the main paper, we further finetune it separately on every test object using the reference images of the object as the training set. The performance of finetuned DeepIM model is shown in Table~\ref{tab:ft_deepim}, which shows that finetuning DeepIM brings significant improvements but still underperforms the Gen6D model which does not train on the object.

\begin{table}[]
    \centering
    \caption{Performance on the GenMOP dataset. ``General" means generalizable or not. ``DeepIM~\cite{li2018deepim}-Ft" means we finetune DeepIM models on every object's reference images separately.}
    \begin{tabular}{C{0.15\linewidth} lccccccc}
    \toprule
    \multirow{2}{*}{Metrics} & \multirow{2}{*}{Method} &  
    \multirow{2}{*}{General} &
    \multicolumn{5}{c}{Object Name} & \multirow{2}{*}{avg.}\\
    & & & Chair & PlugEN & Piggy & Scissors & TFormer & \\
    \midrule
    \multirow{3}{*}{ADD-0.1d}
    &DeepIM~\cite{li2018deepim} &\cmark & 
    12.50 & 6.54 & 29.15 & 18.10 & 31.35 & 19.53\\
    &DeepIM~\cite{li2018deepim}-Ft &\xmark & 
    \textbf{65.50} & 36.92 & 62.31 & 19.40 & 38.49 & 44.52\\
    &Ours   &\cmark &
    61.50& \textbf{19.63}& \textbf{75.38}& \textbf{32.76}&\textbf{62.70}& \textbf{50.39}\\
    \midrule
    \multirow{3}{*}{Prj-5}
    &DeepIM &\cmark & 
    4.50 & 52.34 & 18.50 & 61.64 & 73.81 & 42.16\\
    &DeepIM~\cite{li2018deepim}-Ft &\xmark & 
    40.00 & 70.56 & 82.37 & 73.28 & 98.41 & 72.92\\
    &Ours &\cmark & \textbf{55.00}& \textbf{72.90}& \textbf{92.96}& \textbf{93.53}&\textbf{98.81}& \textbf{82.64}\\
    \bottomrule
    \end{tabular}
    \label{tab:ft_deepim}
\end{table}

\subsection{Fewer reference images}
To show the performance of Gen6D estimator with less reference images, we reduce the reference images on the GenMOP dataset from 128 to 8 by farthest point sampling (FPS) as shown in Table~\ref{tab:view_num}. Sampling 128 images by FPS even slightly improves the performance because FPS makes the view distribute evenly. With 64 reference images, the Gen6D estimator still produces similar results as with all images. When only 16 or 8 reference images retain, the performance reduces reasonably. We further provide detailed Prj-5 of different objects on the GenMOP dataset in Table~\ref{tab:view_num_detail}. This show that the suitable view numbers for different objects are different. 16 reference views on ``piggy" still produce 91\% Prj-5 while 16 views on ``chair" will reduce Prj-5 to 34\%.

\begin{table}[]
    \centering
    \caption{Performance of the Gen6D estimator on the GenMOP dataset with different numbers of reference images. Two metrics are averaged among all objects. ``All" means using all images ($\sim$ 200) from the sequence as reference images.}
    \begin{tabular}{cC{0.08\linewidth} C{0.08\linewidth} C{0.08\linewidth}C{0.08\linewidth}C{0.08\linewidth}C{0.08\linewidth}}
    \toprule
    \multirow{2}{*}{Metrics} & \multicolumn{6}{c}{Reference image number} \\
    & All & 128 & 64 & 32 & 16 & 8 \\
    \midrule
    ADD-0.1d &
    50.39 & 51.30 & 49.68 & 39.45 & 33.56 & 27.55 \\
    Prj-5 
    & 82.64 & 83.28 & 82.30 & 81.03 & 73.51 & 37.33 \\
    \bottomrule
    \end{tabular}
    \label{tab:view_num}
\end{table}
\begin{table}[]
    \centering
    \caption{Prj-5 on GenMOP dataset of Gen6D with different reference image numbers.}
    \begin{tabular}{ccccccc}
    \toprule 
    \multirow{2}{*}{\# Ref. Img.} &
    \multicolumn{5}{c}{Object Name} & \multirow{2}{*}{avg.}\\
        & Chair & PlugEN & Piggy & Scissors & TFormer & \\
    \midrule
    All & 55.00 & 72.90  & 92.96 & 93.53    & 98.81 & 82.64 \\
    128 & 58.00 & 69.63  & 96.48 & 92.67    & 99.60 & 83.28 \\
    16  & 34.00 & 69.63  & 90.95 & 79.74    & 93.25 & 73.51 \\
    8   & 11.00 & 61.68  & 28.14 & 26.29    & 59.52 & 37.33 \\
    \bottomrule
    \end{tabular}
    \label{tab:view_num_detail}
\end{table}

\subsection{Uneven reference image distribution}

\label{sec:view_dist}
\begin{figure}
    \centering
    \includegraphics[width=0.8\textwidth]{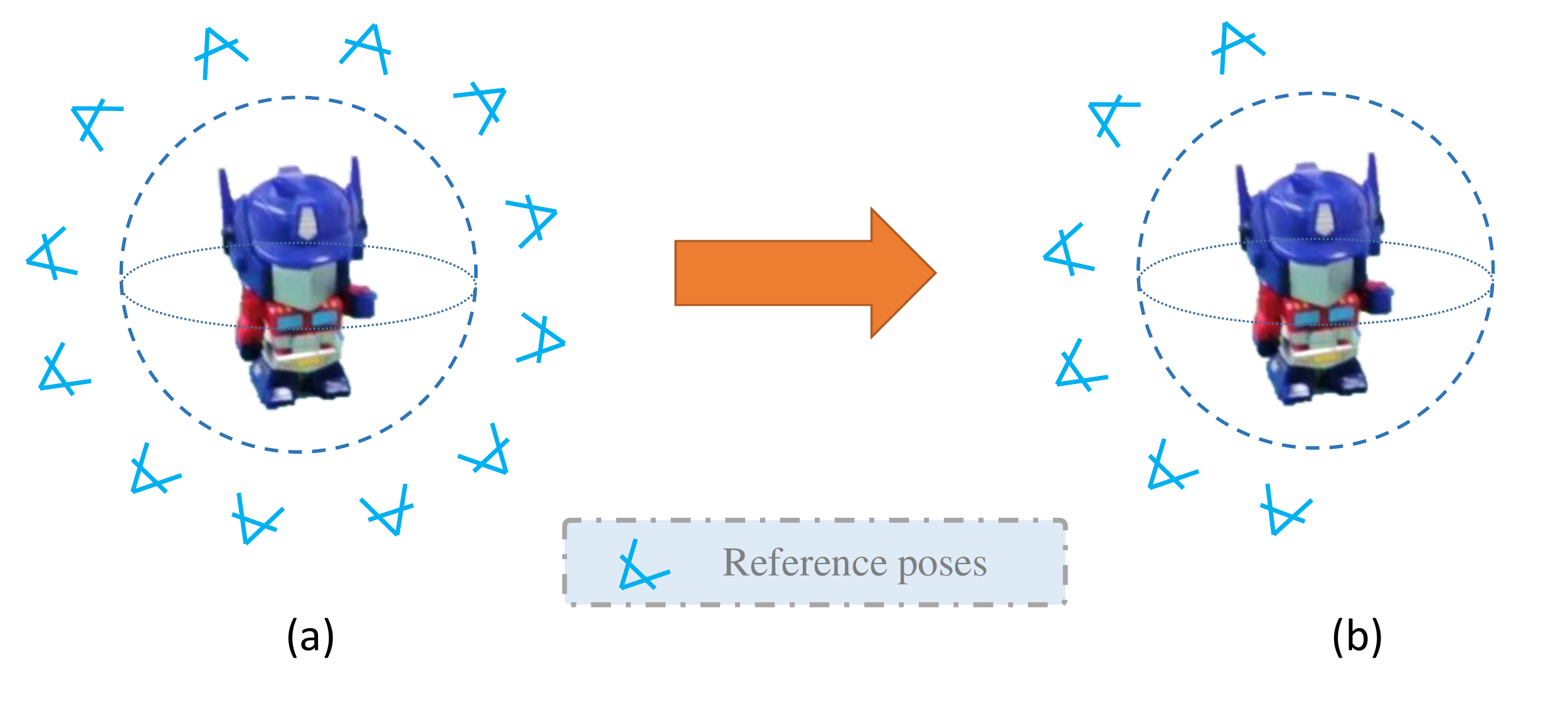}
    \caption{(a) Reference images are homogeneously-distributed around the object. (b) Reference images only distribute in the Y- half space.}
    \label{fig:part}
\end{figure}

By default, reference images are preferred to be distributed evenly around the object as shown in Fig.~\ref{fig:part} (a), which benefits the viewpoint selection and the refinement using neighboring views. When only a part of viewpoints are available like Fig.~\ref{fig:part} (b), Gen6D is not able to accurately predict object poses on images captured from uncovered region. We show the performance of Gen6D using reference images of Fig.~\ref{fig:part} (b) in Table~\ref{tab:uneven}.

\begin{table}[]
    \centering
    \caption{Performance of Gen6D on the GenMOP dataset using homogeneously-distributed reference images (``Even") or only images whose camera centers are in the Y- space (Fig.~\ref{fig:part}(b)) (``Partial").}
    \begin{tabular}{cccccccc}
    \toprule
    \multirow{2}{*}{Metrics} & \multirow{2}{*}{Ref. Img.} &
    \multicolumn{5}{c}{Object Name} & \multirow{2}{*}{avg.}\\
    & & Chair & PlugEN & Piggy & Scissors & TFormer & \\
    \midrule
    \multirow{2}{*}{ADD-0.1d}
    & Even   & 61.50& {19.63}& {75.38}& {32.76}&{62.70}& {50.39}\\
    & Partial & 43.00& {8.53}& {56.28}& {5.17} &{30.95}& {28.79}\\
    \midrule
    \multirow{2}{*}{Prj-5}
    & Even   & {55.00}& {72.90}& {92.96}& {93.53}&{98.81}& {82.64}\\
    & Partial & {32.50}& {7.48}& {73.37}& {17.24}&{47.22}& {35.56}\\
    \bottomrule
    \end{tabular}
    \label{tab:uneven}
\end{table}

\subsection{Refinement iterations}

In Table~\ref{tab:ref_num}, we show results on the GenMOP dataset with different refinement iterations. The results show that applying 1 refinement iteration already greatly improves performance from 17.90 to 38.59 on the ADD-0.1d metric. Further applying 2 refinement iteration will continuously improve the results while using 3 iterations does not. 

\begin{table}[]
    \centering
    \caption{Results of our Gen6D estimator on the GenMOP dataset. ``\#Refine" means the number of refinement iterations used to produce the results.}
    \begin{tabular}{C{0.15\linewidth} ccccccc}
    \toprule
    \multirow{2}{*}{Metrics} &  
    \multirow{2}{*}{\#Refine} &
    \multicolumn{5}{c}{Object Name} & \multirow{2}{*}{avg.}\\
    & & Chair & PlugEN & Piggy & Scissors & TFormer & \\
    \midrule
    \multirow{4}{*}{ADD-0.1d}
    &0      & 14.00& 7.48&39.70& 16.81& 11.51& 17.90\\
    & 1     & 50.50& 9.81& 55.28& 24.57& 52.78& 38.59\\
    & 2     & \textbf{62.00}& \textbf{29.91}& \textbf{80.90}& \textbf{37.07}&56.75& \textbf{53.32}\\
    & 3     & 61.50& 19.63& 75.38& 32.76&\textbf{62.70}& 50.39\\
    \midrule
    \multirow{4}{*}{Prj-5}
    & 0     & 11.50& 40.65& 33.17& 34.05& 64.29& 36.73\\
    & 1     & 44.00& 71.03& 92.96& 84.48& 95.24& 77.54\\
    & 2     & 51.50& \textbf{72.90}& \textbf{94.97}& \textbf{94.40}&\textbf{99.60}& \textbf{82.67}\\
    &3       & \textbf{55.00}& \textbf{72.90}& 92.96& 93.53&98.81& 82.64\\
    \bottomrule
    \end{tabular}
    \label{tab:ref_num}
\end{table}

\subsection{Ablations on training data}
\label{sec:ab_data}
\begin{table}[]
    \centering
    \caption{Performance of Gen6D on the GenMOP and LINEMOD~\cite{hinterstoisser2012linemod} datasets with different training sets. ``Syn." means the ShapeNet~\cite{chang2015shapenet} and Google Scanned Objects~\cite{wang2021ibrnet} datasets. ``GMP" means 5 training objects from the GenMOP dataset. ``LM" means 5 training objects from the LINEMOD~\cite{hinterstoisser2012linemod} dataset. All test objects are not in the training set.}
    \begin{tabular}{C{0.15\linewidth} lccccccc}
    \toprule
    \multirow{2}{*}{Metrics} & 
    \multirow{2}{*}{Trainset} &
    \multicolumn{5}{c}{Object Name} & \multirow{2}{*}{avg.}\\
    
    & & Chair & PlugEN & Piggy & Scissors & TFormer & \\
    \midrule
    \multirow{3}{*}{ADD-0.1d}
     & Syn.+GMP+LM &
    \textbf{61.50}& \textbf{19.63} & \textbf{75.38} & 32.76&\textbf{62.70}& \textbf{50.39}\\
     & Syn.+GMP &
    39.00 & 11.21 & 71.86 & \textbf{37.93}& 39.68 & 39.94\\
     & Syn. &
    30.00 & 21.96 & 61.81 & 24.57& 30.16 & 33.70\\
    \midrule
    \multirow{3}{*}{Prj-5}
     & Syn.+GMP+LM &
    \textbf{55.00}& 72.90& \textbf{92.96}& \textbf{93.53} & \textbf{98.81}& \textbf{82.64}\\
     & Syn.+GMP &
    34.50 & \textbf{77.57} & 92.46 & 91.38 & 89.68 & 77.18\\
     & Syn. &
    10.05 & \textbf{77.57} & 68.84 & 74.57 & 98.41 & 65.98\\
    \midrule
    Metrics & Trainset &  cat  & duck & bvise &  cam  & driller &  avg. \\
    \midrule
    \multirow{3}{*}{ADD-0.1d}
    &Syn.+GMP+LM & \textbf{60.68} & \textbf{40.47} & \textbf{77.03} & \textbf{66.67} & \textbf{67.39} & \textbf{62.45} \\
    &Syn.+GMP & 40.92 & 16.24 & 62.11 & 45.59 & 48.76 & 42.72 \\
    &Syn.     & 31.04 & 11.64 & 61.63 & 35.39 & 54.61 & 38.86 \\
    
    \bottomrule
    \end{tabular}
    \label{tab:ab_data_genmop}
\end{table}

To show the effects of different training data, we show the performance of Gen6D using different training sets in Table~\ref{tab:ab_data_genmop}. Training only on synthetic datasets suffers from the domain gap between the real data and synthetic data. Using real data for training greatly improves the results. LINEMOD~\cite{hinterstoisser2012linemod} brings more obvious improvements than adding GenMOP. The main reason is that LINEMOD has more images ($\sim$1200) on every object while GenMOP only has $\sim 200$ reference images and $\sim 200$ query images for training.

\subsection{Symmetric objects}
\begin{figure}
    \centering
    \begin{tabular}{cccc}
    \includegraphics[width=0.24\textwidth]{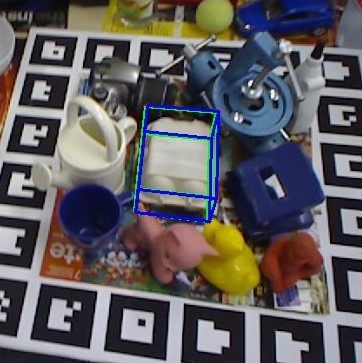} &
    \includegraphics[width=0.24\textwidth]{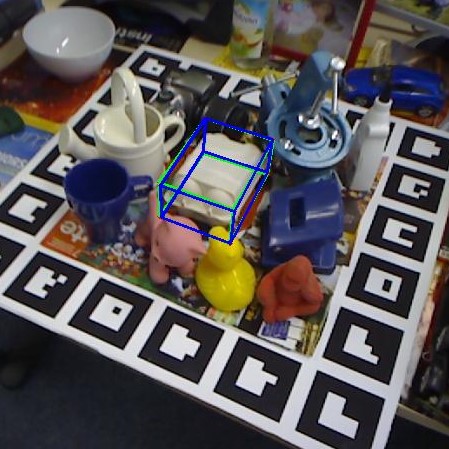} &
    \includegraphics[width=0.24\textwidth]{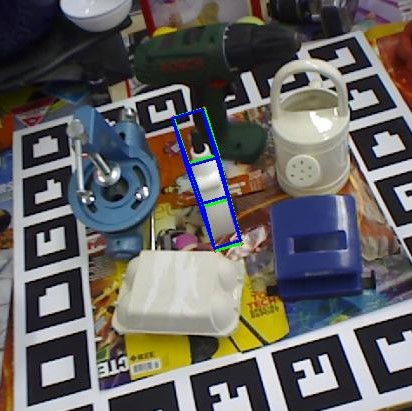} &
    \includegraphics[width=0.24\textwidth]{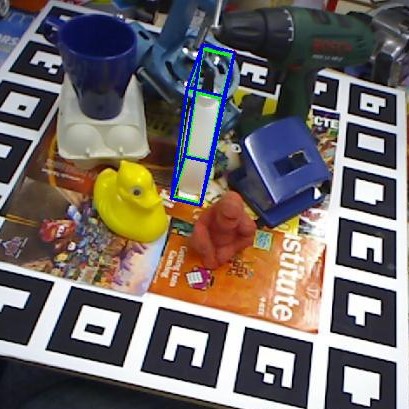}
    \end{tabular}
    \caption{Qualitative results on symmetric objects.}
    \label{fig:sym}
\end{figure}
\begin{table}
    \centering
    \caption{Results on symmetric objects of the LINEMOD~\cite{hinterstoisser2012linemod} dataset. For the ADD metric, we report ``ADD-S-0.1d" which computes the nearest distance between the object points transformed by ground-truth pose and the estimated pose. Note PVNet~\cite{peng2019pvnet} is trained on the specific test object with both synthetic and real images while our Gen6D is not trained on the test object.}
    \begin{tabular}{ccccc}
    \toprule
    \multirow{2}{*}{Name} & \multicolumn{2}{c}{Prj-5} & \multicolumn{2}{c}{ADD-S-0.1d} \\
          & PVNet~\cite{peng2019pvnet} & Ours & PVNet~\cite{peng2019pvnet} & Ours \\
          \midrule
         Eggbox & 99.34 & 97.84 & 99.15 & 98.40\\
         Glue   & 98.45 & 96.24 & 95.66 & 87.16\\
         \bottomrule
    \end{tabular}
    \label{tab:sym}
\end{table}
Gen6D is able to predict poses for symmetric objects. We evaluate Gen6D on two unseen symmetric objects from LINEMOD~\cite{hinterstoisser2012linemod}, i.e. ``glue" and ``eggbox". Qualitative results are shown in Fig.~\ref{fig:sym} and quantitative results are shown in Table~\ref{tab:sym}. The results show that our method is able to achieve reasonable performance on symmetric objects. The main reason is that Gen6D is based on matching query images with reference images. Though symmetry makes multiple feasible poses for a query image, the selector and refiner of Gen6D are able to find reference images near to one of such feasible poses.

\begin{wrapfigure}{r}{0.6\textwidth}
    \centering
    \includegraphics[width=0.59\textwidth]{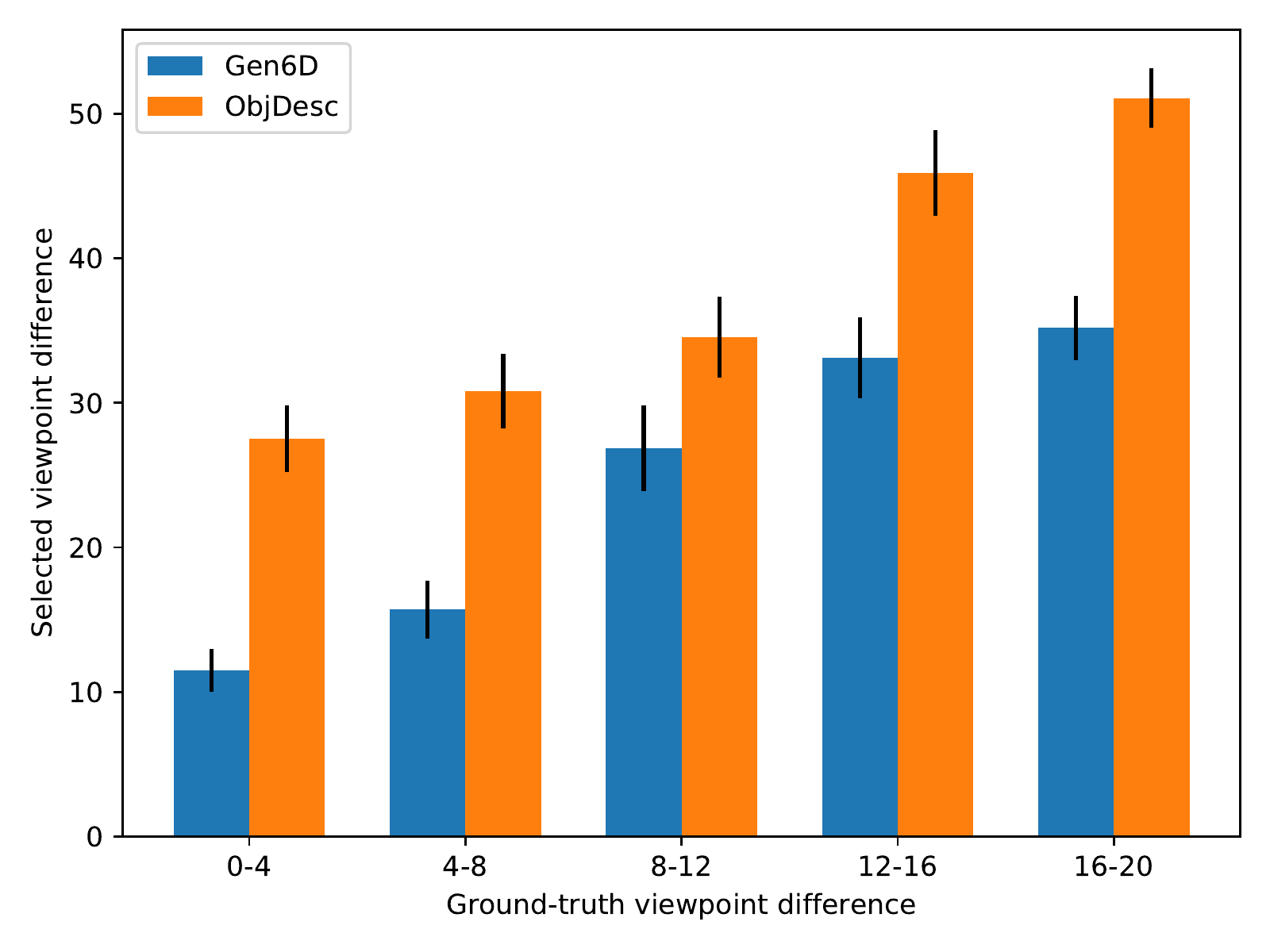}
    \caption{Viewpoint difference between the selected reference image and the query image (y-axis); Viewpoint difference between the ground-truth reference image and the query image (x-axis).}
    \label{fig:view_diff}
\end{wrapfigure}
\subsection{Imperfect viewpoint selection}
\begin{figure}
    \centering
    \setlength\tabcolsep{1.5pt} 
    \begin{tabular}{cccccccc}
        
         Query & GT & Gen6D & ObjDesc & Query & GT & Gen6D & ObjDesc \\
         \includegraphics[width=0.1\textwidth]{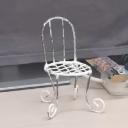}&
         \includegraphics[width=0.1\textwidth]{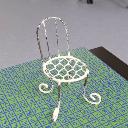}&
         \includegraphics[width=0.1\textwidth]{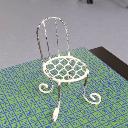}&
         \includegraphics[width=0.1\textwidth]{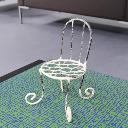} &
         \includegraphics[width=0.1\textwidth]{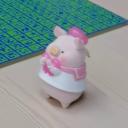}&
         \includegraphics[width=0.1\textwidth]{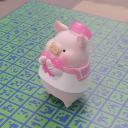}&
         \includegraphics[width=0.1\textwidth]{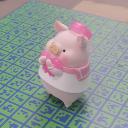}&
         \includegraphics[width=0.1\textwidth]{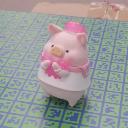}\\
         \includegraphics[width=0.1\textwidth]{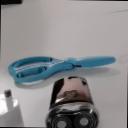}&
         \includegraphics[width=0.1\textwidth]{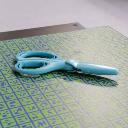}&
         \includegraphics[width=0.1\textwidth]{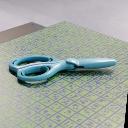}&
         \includegraphics[width=0.1\textwidth]{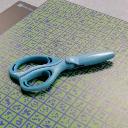}&
         \includegraphics[width=0.1\textwidth]{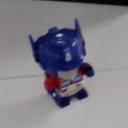}&
         \includegraphics[width=0.1\textwidth]{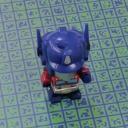}&
         \includegraphics[width=0.1\textwidth]{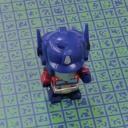}&
         \includegraphics[width=0.1\textwidth]{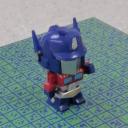}\\
    \end{tabular}
    \caption{The input query image, the ground-truth reference image with nearest viewpoint, the reference images selected by Gen6D and ObjDesc~\cite{wohlhart2015learning}.}
    \label{fig:view_diff_example}
\end{figure}

As discussed in the Section 1 in the main paper, it would be challenging for selector to select the most similar reference image when there is no reference image with an exactly same viewpoint as the query image. To show this, we show Fig.~\ref{fig:view_diff}, where the x-axis shows the viewpoint difference between the ground-truth nearest reference image and the query image while the y-axis shows the viewpoint difference between the selected reference image and the query image. The viewpoint difference is computed as the angle between the query viewpoint and the reference viewpoint by $\arccos \tilde{u}\cdot \tilde{v}$. With the increase of viewpoint difference between the ground-truth reference image and the query image, the viewpoint difference between the selected reference image and the query image is also increasing. However, the proposed selector is able to select more accurate reference image than the baseline ObjDesc~\cite{wohlhart2015learning}. Figure~\ref{fig:view_diff_example} also shows some examples.


\end{document}